\pdfoutput=1

\documentclass[11pt]{article}

\usepackage[]{acl}

\usepackage{times}
\usepackage{latexsym}

\usepackage[T1]{fontenc}

\usepackage[utf8]{inputenc}

\usepackage{microtype}

\usepackage{url}
\usepackage{booktabs}
\usepackage{amsmath}
\usepackage{amssymb}
\usepackage{placeins}
\usepackage{caption}
\usepackage{subcaption}
\usepackage{tikz}
\usepackage{tikz-dependency}

\newcommand{\cready}[1]{}
\usepackage{graphicx}

\usepackage{listings,xfp}
\usepackage{anyfontsize}
\usepackage{xr}

\makeatletter
\newcommand*{\addFileDependency}[1]{
  \typeout{(#1)}
  \@addtofilelist{#1}
  \IfFileExists{#1}{}{\typeout{No file #1.}}
}

{\small 
\xdef\f@size@small{\f@size}
\xdef\f@baselineskip@small{\f@baselineskip}
\normalsize 
\xdef\f@size@normalsize{\f@size}
\xdef\f@baselineskip@normalsize{\f@baselineskip}
}
\newcommand{\smalltonormalsize}{%
  \fontsize
    {\fpeval{(\f@size@small+\f@size@normalsize)/2}}
    {\fpeval{(\f@baselineskip@small+\f@baselineskip@normalsize)/2}}%
  \selectfont
}
\makeatother

\newcommand*{\myexternaldocument}[1]{
    \externaldocument[sup:]{#1}
    \addFileDependency{#1.tex}
    \addFileDependency{#1.aux}
}

\DeclareMathOperator{\dir}{dir}
\DeclareMathOperator{\lab}{lab}

\renewcommand{\vec}[1]{\mathbf{#1}}

\myexternaldocument{emnlp_supp}
\graphicspath{{./graphs/}}

\usepackage{newfloat}
\DeclareFloatingEnvironment[
  fileext = loe,
  listname = Examples,
  name = Example,
  placement = thbp,
  within = none,
  ]{example}
\title{Enhancing the Transformer Decoder with Transition-based Syntax}

\author{ 
  Leshem Choshen\\
  Department of Computer Science\\
  Hebrew University of Jerusalem\\
  {\tt\small leshem.choshen@mail.huji.ac.il}\\
  \And
  Omri Abend \\
  Department of Computer Science\\
  Hebrew University of Jerusalem\\
  {\tt\small omri.abend@mail.huji.ac.il}\\
}
\date{}

\begin{document}
\maketitle

\begin{abstract}
    Notwithstanding recent advances, syntactic generalization remains a challenge for text decoders.
    While some studies showed gains from incorporating source-side symbolic syntactic and semantic structure into text generation Transformers, very little work addressed the decoding of such structure. 
    We propose a general approach for tree decoding using a transition-based approach.
    Examining the challenging test case of incorporating Universal Dependencies syntax into machine translation, we present substantial improvements on test sets that focus on syntactic generalization, while presenting improved or comparable performance
    on standard MT benchmarks. Further qualitative analysis addresses cases where syntactic generalization in the vanilla Transformer decoder is inadequate and demonstrates the advantages afforded by integrating syntactic information.\footnote{Code can be found in \url{https://github.com/borgr/nematus/tree/generation}}
\end{abstract}

\section{Introduction} 

In parallel to the impressive achievements of large neural networks in a variety of NLP fields, more and more work emphasizes the importance of the inductive biases models possess and the types of generalizations they make \citep{welleck2021symbolic,Csordas2021TheDI,Ontanon2021MakingTS}.  Syntactic generalization has been repeatedly identified as a problem in text generation \citep{linzen2020syntactic,hu2020systematic}, an issue that we address here. Importantly, language models may fail, sometimes unexpectedly, on constructions that can be reliably parsed using standard syntactic parsers. 
In this work, we propose a method for incorporating syntax into the decoder to assist in mitigating these challenges, focusing on NMT as a test case.


The use of (mostly syntactic) structure in machine translation dates back to the early days of the field \citep{Lopez:08}. 
While focus has shifted to string-to-string methods since the introduction of neural methods, considerable work has shown gains from integrating linguistic structure into NMT and text generation technologies. We briefly survey such methods in \S\ref{sec:rel_work}.

Incorporating target-side syntax has been less frequently addressed than source-side syntax, possibly due to the additional conceptual and technical complexity it entails, as it requires to jointly generate the translation and its syntactic structure. In addition to linearizing the structure into a string, that allows to easily incorporate source and target structure \citep{Aharoni2017TowardsSN,Nadejde2017PredictingTL}, several works generated the nodes of the syntactic tree using RNNs \citep{gu-etal-2018-top,wang2018tree, wu-etal-2017-sequence}.
Others
have shown gains from multi-task training of a decoder with a syntactic parser \citep{eriguchi2016tree}. However, we are not aware of any Transformer-based architecture to support the integration of target-side structure in the form of a tree or a graph. Addressing this gap, we propose a flexible architecture for integrating graphs  into a Transformer decoder.

Our approach is based on predicting the output tree as a sequence of transitions (\S\ref{sec:trans}), following the transition-based tradition in parsing \citep[][and much subsequent work]{Nivre2003AnEA}. 
The method (presented in \S\ref{sec:synth-arch}) is based on generating the structure incrementally, as a sequence of transitions, as is customary in transition-based parsers. However, unlike standard linearization approaches, our proposed decoder re-encodes the intermediate graph (and not only the generated tokens), thus allowing the decoder to take advantage of the hitherto produced structure in its further predictions. 

In \S\ref{sec:cond}, we discuss the possibilities offered by such decoders, that do not only auto-regress on their previous outputs, but also on (symbolic) structures defined by those outputs. 
Indeed, a decoder thus built can condition both on information it did not predict (e.g., external knowledge bases) and information predicted later on. 
We introduce \emph{bidirectional attention} into the decoder, that allows token representations to encode the following tokens that were predicted. This is similar to the bidirectional attention in the encoder, where any token can attend to any token, and not only to preceding ones.

Our architecture is flexible, supporting decoding not only into trees, but into any graph structure for which a transition system exists.
We test two architectures for incorporating the syntactic graph. One inputs the graph into a Graph Convolutional Network \citep[GCN;][]{kipf2016semi}, and another dedicates an attention head to point at the syntactic parent of each token, which does not yield any increase in the number of parameters.

We assess in \S\ref{sec:experiments} the impact of the proposed architecture on syntactically challenging translation cases \citep{choshen-abend-2019-automatically} and in general. We experiment with a 4 layered model in three target languages, and a 6 layered on En-De. Due to the high computational cost, we experiment with the model on a single language pair only. 
We find that on the syntactic challenge sets proposed by \citet{choshen-abend-2019-automatically}, the proposed decoder achieves substantial improvements over the vanilla decoder, which do not diminish (and even slightly improve) when increasing the size of the model.
In addition, evaluating on the standard MT benchmarks, we find that the syntactic decoders outperform the vanilla Transformer for the smaller model size on all examined language pairs: on the English-German (En-De) and German-English (De-En) challenge sets and on En-De, De-En and English-Russian (En-Ru) test sets, and obtain comparable results to the vanilla when experimenting with a larger model on En-De.
Finally, we analyse the different modifications in isolation, finding that the ablated versions' performance resides between the full model and the vanilla decoder.

\begin{figure*}[t]
	\includegraphics[width=1\linewidth]{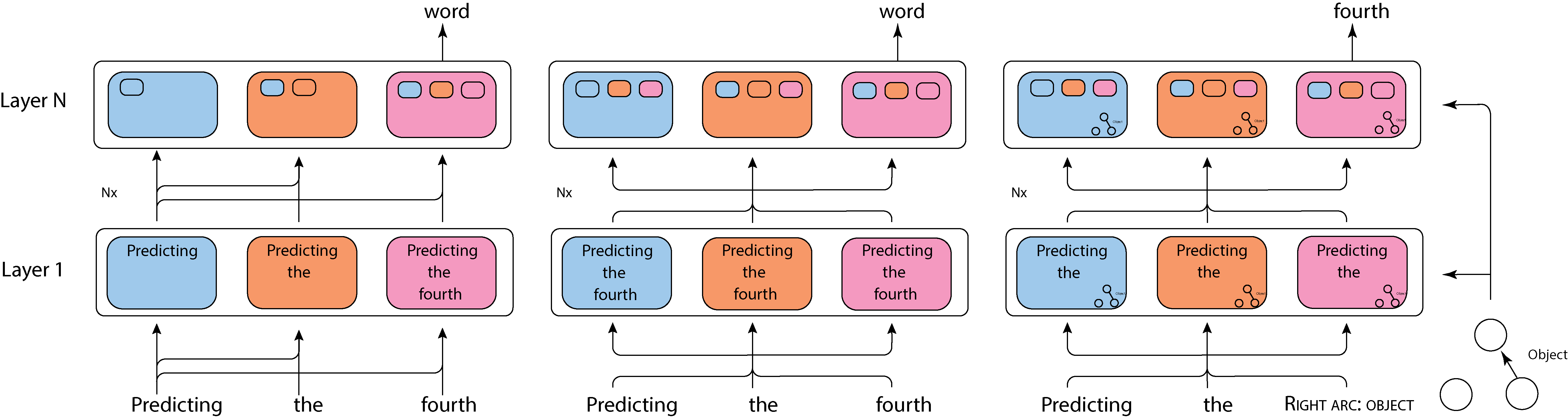}
	\caption{Illustration of the information fed into the decoder with each method. Left: Vanilla. Center: Bidirectional Decoder Right: Structural Decoder. At a given step Bidirectional Decoder attends to all predicted words and Syntactic Transformer predicts edges and receives both edges and words as input.}
	\label{fig:illust}
\end{figure*}

\vspace{-0.1cm}
\section{Decoding Approach}\label{sec:cond}
\vspace{-0.1cm}
\begin{example}
    \centering
    \begin{dependency}
     	\begin{deptext}[column sep=0.35cm]
    	    $\overline{\text{Jo@@~hn}}$ \& put \& the \& coals \& out\\
 	    \end{deptext}
     \deproot[edge height=1.6cm]{2}{root}
     \depedge[edge height=0.3cm]{2}{1}{nsubj}
     \depedge[edge height=0.3cm]{4}{3}{det}
     \depedge[edge height=0.7cm]{2}{4}{obj}
     \depedge[edge height=1.2cm]{2}{5}{compound:prt}
    \end{dependency}
    \caption{Target-side structure reduces the ambiguity of ``put''.
    De source: ``John löschte die Kohlen'' (lit. John put-out the coals).\label{ex:running}}\vspace{-.3cm}
\end{example}

Disambiguating and connecting distant words is a known challenge in NMT \citep{avramidis2020fine}. In Example~\ref{ex:running} to disambiguate ``put''  as not having the sense ``lay'' but  ``extinguish'', ``out'' must be considered. To achieve this from the autoregressed output, the decoder's representation may need to be re-computed after predicting ``out''. 
We note that while source-side information can potentially be used to disambiguate ``put'', it may still be beneficial to enhance the auto-regressive decoder with disambiguating information.


Current implementations impose an architectural bias, namely, a decoded token's representation may not attend to future tokens.
Transformer models mask attention in the following manner (we did not find any alternative methods):
Token embeddings attend only to previously generated tokens, even when the following tokens are already known. This practice ``ensures that the predictions for position $i$ can depend only on the known outputs at positions less than $i$'' \citep{vaswani2017attention}.

We propose to allow attending to any known token (Fig. \ref{fig:illust}), as done on the encoder side. Due to its 
conceptual resemblance to Bidirectional RNN, we name this \textit{Bidirectional Transformer} or biTran.

Formally, let $o_1 \ldots o_n$ be a hitherto predicted sequence and $d$ max sentence length. Attention is $softmax\left(L + M\right)$ where $L\in\mathbb{R}^{d\times d}$ are the logits and $M\in\mathbb{R}^{d\times d}$ is a mask. Hence, $M(i,j)=-\infty$ masks a token $j$ from representation $i$.

\noindent
\begin{equation*}
\vspace{-0.1cm}
\small
M_v(i,j)=
\begin{cases}
0 &  j<i \\
-\infty &  o.s.
\end{cases}
\vspace{-0.1cm}
\end{equation*}
\noindent
while Bidirectional attention mask is 
\noindent
\begin{equation*}
\vspace{-0.1cm}
\small
M_{bi}(i, j)=
\begin{cases}
0 & n<j \\
-\infty & o.s.
\end{cases}
\vspace{-0.1cm}
\end{equation*}

This change does not introduce any new parameters or hyperparameters, but still increases the expressivity of the model. We note, however, that this modification does prevent some commonly implemented speed-ups relying on unidirectionality \citep[e.g., in NEMATUS;][]{Sennrich2017NematusAT}.


Apart from the technical contribution, we emphasize that this and the following approaches take advantage of attention-based models being state-less. Transformers can, therefore, be viewed as conditional language models, namely as models for producing a distribution for the next word, given the generated prefix and source sentence. Viewing them as such opens possibilities that were not native to RNNs, such as predicting only partial outputs and conditioning on per-token or non-autoregressed context (see App. \ref{sec:conditional}).

\section{Transition-based Structure Generation}\label{sec:trans}
\vspace{-0.2cm}

We turn to describe how we represent structure within the proposed decoder. 

We generate the target-side structure with a transition-based approach, motivated by the practical strength of such methods, as well as their sequential nature, which fits neural decoders well. We therefore augment the vocabulary with transitions.
Our work is inspired by RNNG \citep{Dyer2016RecurrentNN}, a conceptually similar architecture that was developed for RNNs.
At each step, the input to the decoder includes the tokens and the parse graph that was generated thus far.
As edges and their tokens are not generated simultaneously (but rather by different transitions; see below), we rely on bidirectional attention to update the past embeddings when a new edge connects previously generated tokens. In this section, we present the syntactic transitions and in the next (\S\ref{sec:synth-arch}), the ways we incorporate it back into the model.

In this work, we represent syntax through Universal Dependencies \citep[UD;][]{nivre2016universal}, but note that other syntactic and semantic formalisms that have transition-based parsers \citep{hershcovich2018multitask,stanojevic-steedman-2020-max,oepen-etal-2020-mrp} fit the framework as well. We select UD due to its support for over 100 languages and its status as the de facto standard for syntactic representation. 

We base our transition system on arc-standard \citep{Nivre2003AnEA}, which can produce any projective tree. Both contain a transition connecting two words by a labeled edge. However, we replace {\sc Shift} that reads the next word by {\sc subword$_t$} generating a new sub-word $t$. Sub-words are generated successively until a full word is formed. To avoid suboptimal representation of transition tokens, we add the edges going through them to the graph (e.g., the edge {\sc Left-Arc}{\it :det}$\xrightarrow[]{\text{det}} the$).

We denote with $f$ the transition functions updating a word stack $\Sigma$ and the labeled graph $G$. If $a,b$ are the top and second words in $\Sigma$ respectively, and $x$ a transition, then $f(x;\Sigma)$ is defined as:

\begin{table}[h]
\vspace{-0.15cm}
\centering
\small
\begin{tabular}{lcc}
x (token) & $\Sigma$ & Edges Added \\
\hline
$ Subword_t $ & t,a,b & $\emptyset$ \\
\textsc{Left-Arc}{\it :l} & a & $ a \xrightarrow[]{\text{l}} b$, $ x \xrightarrow[]{\text{l}} b$, $ a \xrightarrow[]{\text{l}} x $\\
\textsc{Right-Arc}{\it :l} & b & $ b \xrightarrow[]{\text{l}} a $, $ x \xrightarrow[]{\text{l}} a$, $ b \xrightarrow[]{\text{l}} x$
\end{tabular}
\vspace{-0.15cm}
\end{table}


For brevity, we denote an edge from/to every subword of $a$ as an edge from/to $a$.
Overall, the translation sequence to create the graph in Example \ref{ex:running} is: Jo@@ hn put {\sc Left-Arc}{\it :nsubj} the coals {\sc Left-Arc}{\it :det}
{\sc Right-Arc}{\it :obj}  out {\sc Right-Arc}{\it :compound:prt} (more details in App. \ref{sec:setup})
 


     



\section{Regressing on Generated Structure} \label{sec:synth-arch}

As discussed in \S\ref{sec:cond}, the state-less nature of the Transformer allows re-encoding not only the previous predictions, but any information that can be computed based on them. 
So far, we proposed to autoregress on the syntactic structure, token by token. However, as $f$ is deterministic, learning to emulate it, is pointless. Instead, we can autoregress on the generated graph itself, $G=f\left( o_1\ldots o_n\right)$, as well as the encoder output, $o_1\ldots o_n$.

Our approach is modular and works with any graph encoding method. 
We experiment with two prominent methods for source-side graph encoding.

\paragraph{GCN Encoder.}
Graph Convolutional Networks \cite[GCN;][]{kipf2016semi} are a type of graph neural network. GCNs were used  successfully by previous work to encode source-side syntactic and semantic structure for NMT \citep{bastings-etal-2017-graph,marcheggiani-etal-2018-exploiting}.
The GCN layers are stacked immediately above the embedding layer.
The GCN contains weights per edge type and label as well as gates, that allow placing less emphasis on the syntactic cue if the network so chooses. Gating is assumed to help against noisy structure, which machine output is expected to be. See ablation experiments to assess the impact of gating in \S\ref{sec:ablat}.

Following \citet{kipf2016semi}, we introduce 3 edge types. \textit{Self} from a token to itself, \textit{Left} to the parent tokens and \textit{Right} from the parents.

A GCN layer over input layer $h$, a node $v$ and a graph $G$ containing nodes of size $d$, with activation $\rho$, edge directions $\dir$, labels $\lab$, and a function $N$ from a node in $G$ to its neighbors is

\noindent
\begin{equation*}
\small
	\vec{gcn}(h, v, G) = \rho \Bigg( \sum_{u \in \mathcal{N}(v)} g_{u, v}\cdot f_{u,v} \Bigg)
\end{equation*}

\noindent
where $f_{u,v}$ are graph weighted embedding:

\noindent
\begin{equation*}
\small
f_{u, v}=\left( W_{\dir(u, v)} \, \vec{h}_u + \vec{b}_{\lab(u, v)} \right)
\end{equation*}

\noindent
and $g_{u, v}$ is the applied gate:

\noindent
\begin{equation*}
\small
	g_{u, v}=\sigma \Big( \vec{h}_u \cdot \vec{\hat{w}}_{\dir(u,v)} + \hat{b}_{\lab(u, v)} \Big)
\end{equation*}

\noindent where $\sigma$ is the logistic sigmoid function and $\vec{\hat{w}}_{\dir(u,v)} \in \mathbb{R}^d$, $W \in \mathbb{R}^{d \times d}$, $\hat{b}_{\lab(u, v)} \in \mathbb{R}$, $\vec{b} \in \mathbb{R}^d$ are the learned parameters for the GCN. 

\paragraph{Attending to Parent Token.} \label{sec:parent} 
The second re-encoding method we test, {\sc Parent}, dedicates an attention head only to the parent(s) of the given token. Commonly, the parent is given by an external parser \citep{hao2019multi} or learned locally in each layer, to focus the attention \citep{strubell-etal-2018-linguistically}.
Unlike such approaches, we define the parents by the self-generated graph. To allow ignoring it when preferable or when no parent was generated, we also allow attending to the current token. To recap, for a token $o_i$, we mask all but $o_i$ and its parents.

{\sc Parent} differs from GCN considerably. On the one hand, {\sc Parent} requires minimal architectural changes and no additional hyperparameters. It also affects different network parts, some attention heads, rather than an additional embedding. On the other hand, only GCN represents the labels and the whole graph, specifically children. By considering both architectures, we show that graph methods for the encoder \citep{bastings-etal-2017-graph} may be easily adapted to the decoder, demonstrating the flexibility of the proposed framework.

\vspace{-0.2cm}
\section{Experimental Setup}\label{sec:methodology}

\paragraph{Metrics.} We report both BLEU \citep{Papineni2002BleuAM} and chrF+ \citep{Popovic2017chrFWH} and note that chrF+ has been deemed more reliable for current technology \citep{ma2019wmt19metrics}.

\paragraph{Model.} Medium (large) models are trained with batch size 128, embedding size 256 (512), 4 (6) decoder and encoder blocks, 8 attention heads ({\sc Parent} replaces one). 
We train for 90K (150K) steps, where empirically some saturation is reached, allowing a fair system comparison \citep{popel2018training}.
The GCN architecture includes 2 layers with residual connections. Parses are extracted by UDPipe \citep{straka2018udpipe}, UD2.0 for English and German and UD2.5 syntagrus for Russian.

Unable to identify a preexisting implementation, we implemented labeled sparse GCNs with gating in Tensorflow. Implementation mostly focused on memory considerations, and was optimized for runtime when possible. More on implementation details, filtering and preprocessing in App. \ref{sec:setup}.

\paragraph{Language Pairs.} We experiment on 3 language pairs with 3 target languages: English (De-En), German (En-De) and Russian (En-Ru).
We use the WMT16 data \citep{bojar2016findings} for En-De, and either the clean News commentary or the full noisy WMT20 data \cite{Barrault2020FindingsOT} for En-Ru.

\paragraph{Test sets.} 
Newstest 2012 served as a development set.
To measure the overall system performance we used newstest 2013-15. 

To test syntactic generalization, we used the challenge sets by \citet{choshen-abend-2019-automatically}.
Those are sub-sets of the books and newstest corpora in En$\leftrightarrow$De, automatically filtered by a syntactic parser to contain {\it lexical long-distance dependencies}. i.e., sentences where two or more non-consecutive words correspond to a single word. E.g., ``put ... out'' in Example \ref{ex:running} corresponds to the German ``l\"{o}schte'' (see also Example~\ref{tab:challenge_example}). Previous work has shown such phenomena to be challenging for present-day NMT systems.

Improving the automatic measures on one such challenge set indicates better performance on a specific phenomenon, while better overall challenge set performance implies better handling of lexical long-distance dependencies.

The various challenge set settings are represented as a 3-tuple $(dir,p,dom)$, corresponding to the  direction, inspected phenomenon and domain. 
\textit{Direction} can be either ``source'' or ``target'', indicating whether the long distance dependency is in the source or the target (i.e., in the reference). A more effective representation of the target-side syntax should improve target challenges and potentially also the source side's, by increasing the model's ``awareness'' to syntactic structure.
By \textit{phenomenon}, we refer to the syntactic phenomenon in question.  There are three test cases for English phenomena and two for German. 
By \textit{domain} we refer to the origin of the examples, which can be either the sizable {\it books} corpus \citep{Tiedemann2012opus}, or a smaller news corpus \citep{Barrault2020FindingsOT}.



\section{Results}\label{sec:experiments}

\begin{table*}[tbh!]
\small
\centering

\small
\centering

\begin{subtable}[h]{\textwidth}
\resizebox{\textwidth}{!}{%
\begin{tabular}{lrrrrrrrrrrrr} & \multicolumn{4}{c}{Preposition Stranding} & \multicolumn{4}{c}{Particle} & \multicolumn{4}{c}{Reflexive} \\
 & \multicolumn{2}{c}{Books}  & \multicolumn{2}{c}{News} & \multicolumn{2}{c}{Books} & 
 \multicolumn{2}{c}{News} & 
 \multicolumn{2}{c}{Books}  & 
 \multicolumn{2}{c}{News} \\
 & BLEU & chrF+ & BLEU & chrF+ & BLEU & chrF+ & BLEU & chrF+ & BLEU & chrF+ & BLEU & chrF+ \\
Vanilla & -- & -- & -- & -- & 4.14 & 20.72 & 20.31 & 49.04 & 8.08 & 32.38 & 20.65 & 49.09 \\
{\sc Parent} & -- & -- & -- & -- & \textbf{8.37} & \textbf{33.78} & \textbf{20.54} & \textbf{49.99} & \textbf{8.60} & \textbf{33.49} & \textbf{21.39} & \textbf{50.01}

\end{tabular}%
}
\caption{Target challenge sets for En-De, large models (Preposition Stranding is omitted as it is not present  in German)\label{tab:trg_large_syn}}
\end{subtable}

\begin{subtable}[h]{\textwidth}
\resizebox{\textwidth}{!}{%
\begin{tabular}{lrrrrrrrrrrrr}
Vanilla & 8.70 & 33.58 & \textbf{13.82} & 43.41 & 8.59 & 32.66 & \textbf{15.28} & 44.28 & 8.54 & 32.85 & 18.90 & 45.82 \\
{\sc Parent} & \textbf{9.03} & \textbf{34.83} & 11.53 & \textbf{45.12} & \textbf{8.59} & \textbf{33.71} & 14.99 & \textbf{45.90} & \textbf{9.05} & \textbf{34.11} & \textbf{20.79} & \textbf{46.73}
\end{tabular}%
 }
\caption{Source challenge sets for En-De, large models\label{tab:large_syn}}
\end{subtable}

\begin{subtable}[h]{\textwidth}
\resizebox{\textwidth}{!}{%
\begin{tabular}{lrrrrrrrrrrrr}
Vanilla & 5.95 & 25.88 & 9.96 & 36.96 & 5.37 & 24.69 & 9.39 & 39.19 & 5.32 & 24.71 & \textbf{16.48} & 42.04 \\
{\sc Parent} & \textbf{6.21} & \textbf{28.12} & 11.17 & \textbf{41.13} & 5.47 & \textbf{25.74} & \textbf{11.93} & \textbf{41.24} & \textbf{5.71} & \textbf{26.22} & 15.56 & 42.76 \\
GCN & \textbf{6.21} & 27.27 & \textbf{11.31} & 40.48 & \textbf{5.51} & 25.53 & 10.35 & 39.83 & 5.46 & 25.70 & 16.45 & \textbf{43.03} \\
\end{tabular}%
}
\label{tab:de_en_challenge}
\caption{Source challenge sets for En-De, medium models}
\end{subtable}


\centering
\begin{subtable}{\textwidth}
\centering
\resizebox{\textwidth}{!}{%
\begin{tabular}{lrrrrrrrrrrrr}
Vanilla & 6.38 & 27.30 & 9.18 & 38.22 & 6.53 & 25.70 & 10.54 & 38.28 & 6.15 & 25.94 & 17.20 & 43.12 \\
{\sc Parent} & \textbf{7.59} & \textbf{27.87} & \textbf{10.81} & 39.22 & \textbf{7.07} & \textbf{26.50} & 9.72 & 39.57 & \textbf{6.82} & \textbf{26.58} & 17.56 & 44.00 \\
GCN & 6.33 & 26.60 & 10.14 & \textbf{41.00} & 6.69 & 26.16 & \textbf{10.60} & \textbf{39.81} & 6.33 & 25.83 & \textbf{20.16} & \textbf{44.19} \\
\end{tabular}%
}
\caption{Target challenge sets for De-En, medium models}
\label{tab:trg_en_de_challenge}
\end{subtable}


\caption{Results on the syntactic challenge sets, both on the larger sets from books and the smaller ones from news. Models include Vanilla and the GCN and {\sc Parent} UD-based decoders. Models can be large or medium in size and trained on En-De or De-En. Challenges are either in the source or target translation. See also App. \ref{sec:res}. \label{tab:all_challenge}}
\end{table*}

\begin{example}[tbp]
\smalltonormalsize
\begin{tabular}{lll}
Source &  der gruppe, an die sich der Plan richtet &  \\
Gloss &  the group to which himself the plan aims &  \\\vspace{.15cm}
Ref. &  the group to whom the plan is aimed &  \\
{\sc Parent} &  the group to which the plan is aimed &  \\
Vanilla &  the group aimed at the plan & 
\end{tabular}
\caption{A part of a sentence with a long-distance German reflexive verb from the challenge set.\label{tab:challenge_example} \vspace{-0.3cm}}
\end{example}

We compare the syntactic generalization abilities of the different decoders in \S\ref{sec:synt_res}, and continue by examining their overall performance (\S\ref{sec:overall}). 
We then assess the contribution of the components of the system through ablation experiments (\S\ref{sec:ablat}) and evaluate the effects of noisy training data (\S\ref{sec:noise}).


\subsection{Syntactic Generalization}\label{sec:synt_res}

We evaluate the syntactic generalization abilities of the models using the syntactic challenge sets. Results (Table \ref{tab:all_challenge}) show that the medium {\sc Parent} (GCN) improves over the Vanilla in 18 (20) of 20 target challenge settings and 19 (19) of 20 in the source challenges. The large model improves in 18/20 of the challenges and gains seem similar or even larger. The latter results suggest that simply using larger models is unlikely to address these gaps in syntactic generalization. See also \ref{sec:res}.



\subsection{Overall Performance}\label{sec:overall}

\begin{table*}[tbh!]
\small
\centering
\begin{subtable}{\textwidth}
\centering
\begin{tabular}{lrr|rr|rr}
 & \multicolumn{2}{c}{2013} & \multicolumn{2}{c}{2014} & \multicolumn{2}{c}{2015} 
 \\
 & \multicolumn{1}{c}{BLEU} & \multicolumn{1}{c}{chrF+} & \multicolumn{1}{c}{BLEU} & \multicolumn{1}{c}{chrF+} & \multicolumn{1}{c}{BLEU} & \multicolumn{1}{c}{chrF+} 
 \\
Vanilla & 17.61 & 45.54 & 18.23 & 47.29 & 19.57 & 47.50 
\\
{\sc Parent} & \textbf{18.11} & \textbf{46.75} & 18.6 & \textbf{48.46} & \textbf{20.55} & \textbf{49.20} 
\\
GCN & 18.03 & 46.43 & \textbf{18.86} & \textbf{48.46} & 20.32 & 48.90 
\\
\hline
BiTran & 17.64 & 45.66 & 18.34 & 47.53 & 19.33 & 47.61 
\\
Linearized & 17.71 & 46.07 & 18.39 & 47.69 & 19.81 & 48.36 
\\
- Gates & 17.81 & 46.12 & 18.43 & 48.08 & 20.06 & 48.62 
\\
- Labels & 17.98 & 46.40 & 18.77 & 48.29 & 19.96 & 48.73 
\\
\end{tabular}%
\caption{Overall performance for En-De, medium models}
\end{subtable}
\begin{subtable}{\textwidth}
\centering
\begin{tabular}{lrr|rr|rr}

Vanilla & \textbf{23.64} & 53.44 & 21.94 & 53.13 & \textbf{21.60} & \textbf{50.84} 
\\ 
{\sc Parent} & 23.56 & \textbf{54.08} &  \textbf{22.11} & \textbf{53.77} & 20.69 & 49.16 
\end{tabular}%
\caption{Overall performance for En-De translation, large models\label{tab:large}}

\end{subtable}
\begin{subtable}{\textwidth}
\centering
\begin{tabular}{lrr|rr|rr}
Vanilla & 21.51 & 48.20 & 21.4 & 48.46 & 21.44 & 48.13 
\\
{\sc Parent} & \textbf{22.46} & 49.24 & 21.75 & 49.41 & 22.14 & 49.31 
\\
GCN & 22.33 & 49.27 & 21.76 & 49.71 & \textbf{22.43} & \textbf{49.73} 
\\ 
\hline
BiTran & 21.63 & 48.48 & 21.42 & 48.86 & 21.38 & 48.54 
\\
Linearized & 21.95 & 49.27 & 21.83 & \textbf{49.79} & 22.2 & 49.70
\\
- Gates & 22.28 & 49.33 & \textbf{21.89} & 49.68 & 22.04 & 49.39 
\\
- Labels & 22.21 & \textbf{49.46} & 21.75 & 49.73 & 22.26 & 49.57 
\\
\end{tabular}%
\caption{Overall performance for De-En translation, medium models}
\end{subtable}

\begin{subtable}{\textwidth}
\centering
\begin{tabular}{lrr|rr|rr}
Vanilla & 13.2 & 38.72 & 17.17 & 43.69 & 14.19 & 40.87 
\\
{\sc Parent} & \textbf{13.61} & \textbf{40.67} & \textbf{18.53} & \textbf{46.44} & \textbf{15.75} & \textbf{43.57} 
\\
GCN & 13.25 & 40.31 & 17.86 & 46.09 & 15.38 & 43.09 
\end{tabular}%
\caption{Overall performance for En-Ru\label{tab:rus}}
\vspace{-.2cm}
\end{subtable}
\caption{Overall performance in different settings. Ablated models (where applicable), appear in the bottom part of the table and include the Bidirectional Transformer (BiTran), with linearized syntax (Linearized), GCN without labels or gating (-Gates) and GCN without labels (-Labels). The syntactic variants consistently outperform the vanilla and ablated variants in the medium size setting and are comparable to it in the large one. The Bidirectional Transformer (BiTran) slightly outperforms Vanilla Transformer.\label{tab:english-german}}
\end{table*}

Table \ref{tab:english-german} presents the   
overall test performance for all models.
For medium-sized models, the UD-based decoders (GCN and {\sc Parent} rows) show better performance over the vanilla decoder in all settings, with 0.7-1.1 average BLEU improvements and 1-2.4 chrF+. We see a slight advantage to the GCN decoder on De-En, and an advantage to {\sc Parent} on En-De and En-Ru. 
We apply a sign test on all medium size test sets and separately on challenge sets. GCN and {\sc parent} are significantly ($p<0.01$) better than BiTran, which is significantly better than Vanilla Transformer.

With the large models, {\sc parent} performs comparably to the vanilla  (Table \ref{tab:large}), despite the superior results it obtains on syntactic generalization.

\subsection{Ablation Experiments}\label{sec:ablat}

In order to better understand the contribution of different parts of the architecture and to compare them, we consider ablated versions (See Tables \ref{tab:english-german} and App. \ref{sec:res}). Differences are small but consistent. In one, {\it Linearized}, we train the vanilla Transformer over the transitions, linearized to a string, without encoding the graph through GCN or attention. This is reminiscent of the approaches taken by \citet{Aharoni2017TowardsSN, Nadejde2017PredictingTL}, albeit with a different form of linearization. 
Results place Linearized in a clear place: consistently better than the structure-unaware models but not as good as the structure-aware ones.


We turn to experiment with ablated versions of the GCN decoder. \textit{Unlabeled} ignores the labels and relies only on the graph structure, while \textit{Ungated}, also removes the gate $g$. Gating was hypothesized to be important to avoid over-reliance on the erroneous edges \citep{bastings-etal-2017-graph,hao2019multi}. As our graphs are generated by the network, rather than fed into it by an external parser, this is a good place to test this hypothesis.

Comparing GCN with and without labels, we find their contribution to be limited. Despite some improvement in overall BLEU, as often as not, \textit{Unlabeled} is better on the challenges. We advise caution, however, in interpreting these results, as they may not necessarily indicate that syntactic labels are redundant. There are two technical points to consider. First, the labels' role in GCNs is small, they contribute many hyperparameters, while only affecting a bias term. Presumably, this is an inefficient use that should be addressed in future work. Second, the labels are incorporated also through the transitions, and hence have token embeddings. These could compensate for the disregard of labels.

Unlike labels, gating appears to be crucial. The Ungated scores are lower than the Unlabeled variant in 34/40 challenges. This might indirectly support the hypothesis that gating aids with erroneous parses. It also hints introducing similar mechanisms to {\sc Parent} may also be beneficial.

Even BiTran provides a small (up to $.28$ BLEU,$.42$ chrF+) but consistent improvement. Indeed, it outperforms the vanilla on average and in 10/12 scores in each pair.
We observe a similar trend in the challenge sets (Table \ref{tab:all_challenge}): BiTran improves scores in 26/40 syntactic challenge sets. In conclusion, bidirectionality in itself is somewhat beneficial, both in general and specifically for aggregating the syntactically correct context tokens.

As a next step, we compare GCN ablations to {\sc Parent}. Like unlabeled GCNs, {\sc Parent} does not rely on the labels and provides a different way to incorporate the graph structure, which is still shown to be successful. We note that while labels are not incorporated, they appear as transition inputs and can be attended to. 
Comparing the two architectures, {\sc Parent} shows significant gains over Unlabeled GCN. Despite being easier to implement and being much lighter in terms of memory, time and hyperparameters, {\sc Parent} generally outperforms Unlabeled GCN in both performance and specific challenges. {\sc Parent} is slightly better than unablated? GCN on En-De and slightly worse on De-En. It is better on 3 of 5 De-En phenomena and one of the En-De, when compared to the GCN variant.


\subsection{Noise Robustness}\label{sec:noise}


Preliminary experiments indicate that syntactic architectures may be more sensitive to noisy training data than the vanilla Transformer, possibly amplifying parser errors. To test this, we trained on the full WMT data for En-Ru, which is mostly crawled data. Results show that the improvement in chrF+ is smaller, 1 point instead of 1.5-2.5 in other settings, and BLEU scores are somewhat worse (see App.~\S\ref{sec:full_noisy}). It seems then that overall, the inclusion of noisy data diminishes the relative improvement. 

An alternative explanation to these results may be that our methods contribute less in the presence of more training data. Our positive results on En-De and De-En, that use relatively large amounts of data (4.5M sentence pairs), show that if this is indeed the case, saturation is slow.




\subsection{Qualitative Analysis}

To complement the automatic challenges, 
we compile a set of 99 simple subject-verb-object sentences where the German object and subject can swap locations without affecting the meaning. 
We created three sets of sentences, where the case marking for the subject and object may or may not be ambiguous. For example, \textit{Das Pferd bringt der Vater} and \textit{Der Vater bringt das Pferd} both translate to \textit{the father brings the horse}.
Such examples are of particular interest to us here, as the case of the first noun phrase is ambiguous (``Das Pferd'' could be either a subject or an object) and is only disambiguated by the case marking of the second one. These cases require some understanding of the syntax to translate correctly.
See App.~\S\ref{sec:mixup}.

A native-speaking German annotator, fluent in English, then evaluated the medium-size {\sc Parent} and Vanilla outputs on these sentences.
The ambiguous examples were found to be challenging for both systems, especially the ambiguous case markings. However, overall, {\sc Parent} is more robust to the changes in order. 
Interestingly, both models ({\sc Parent} more consistently) translate some sentences to passive voice, keeping both the (changed) order and the meaning.

\section{Related Work}\label{sec:rel_work}	

While there are indications that Transformers implicitly learn some syntactic structure when trained as language models or as NMT \citep[e.g.,][]{jawahar2019bert,manning2020emergent, don2022prequel}, it is not at all clear whether such information replaces the utility of incorporating syntactic structure. Indeed, a considerable body of work suggests the contrary.


Much previous work tested RNN-based and attention-based systems for their ability to make structural generalizations \citep{welleck2021symbolic,Csordas2021TheDI,Ontanon2021MakingTS}. Syntactic generalizations seem to pose a particularly difficult challenge \citep{Ravfogel2019StudyingTI,mccoy2019right}. Moreover, while NMT often succeeds in translating inter-dependent linearly distant words, their performance is unstable: the same systems may well fail on other ``obvious'' cases of the same phenomena \citep{belinkov2018synthetic,choshen-abend-2019-automatically}. This evidence provides motivation for efforts such as ours, to incorporate linguistic knowledge into the architecture.

Syntactic structure was used to improve various tasks, including code generation \citep{Chakraborty2018Tree2TreeNT}, question answering \citep{bogin2020latent}, automatic proof generation \citep{gontier2020measuring} language modelling \citep{wilcox2020structural} and grammatical error correction \citep{Harer2019TreeTransformerAT}.
Such approaches, however, are task specific. E.g., the latter makes strong conditional independence assumptions, and is less suitable for MT where the source and target syntax may diverge considerably. 


In NMT, some works used structural cues by reinforcement learning \citep{Wieting2019BeyondBT, yehudai}, but the gain from such methods seems to be constrained by the performance presented by the pre-trained model \citep{Choshen2020OnTW}.
\citet{aharoni2017morphological} proposed to replace the source and target tokens with a linearized constituency graph. \citet{Nadejde2017PredictingTL} proposed a similar approach using CCG parses. \citet{eriguchi2016tree} proposed an RNN to encode the source syntax. Some works suggested modifications to the RNN to encode source-side syntax \citep{chen-etal-2017-neural,chen2018syntax,li-etal-2017-modeling}.
\citet{song-etal-2019-semantic}  used a graph recurrent network to encode source-side AMR structures.
Few works suggested changes in the Transformer to incorporate source-side syntax: \citet{Nguyen2020TreestructuredAW} and \citet{Bugliarello2020EnhancingMT} proposed a tree-based attention mechanism to encode source syntax; \citet{zhang2019syntaxEnhanced} incorporated the first layers of a parser in addition to the source-side token embeddings.
Relatedly, previous work showed gains from using syntactic information for preprocessing \citep{ponti2018isomorphic,zhou-etal-2019-handling}.

Much fewer works focused on structure-based decoding. \citet{eriguchi2017learningTP}, building on \citet{Dyer2016RecurrentNN},  train a decoder in a multi-task setting of translation and parsing. Notably, unlike in the method we propose, their generated translation is not constrained by the parse during the decoding.
Few works proposed alternating between two connected RNNs one translating and one creating a linearized graph using a tree-based RNN \citep{wang2018tree} or transition-based parsing \citep{ wu-etal-2017-sequence}. \citet{gu-etal-2018-top} both parse and generate, using a recursive RNN representation.


Other work changed RNNs \citep{Tai2015ImprovedSR} or Transformers to include structural inductive biases, but without explicit syntactic information. \citet{Wang2019TreeTI} suggested an unsupervised way to train Transformers that learn tree-like structures following the intuition that such representations are more similar to syntax. \citet{Shiv2019NovelPE} encoded tree-structured data in the positional embeddings.

\section{Discussion}


The work we presented is motivated from several angles. 
First, we note that Transformers are trained in the same way that former sequence to sequence models are trained (e.g., RNNs) and to many, they are just a better architecture for the same task. Instead, our work emphasizes the possibility of conditional training using Transformers; namely, Transformers should be able to predict the third token given the first two, even without previously predicting them. Although generally not implemented this way, Transformers are already conditional networks, and allow for flexibility not found in RNNs. 


The finding that MT quality changes between beginnings and ends of predicted sentences both in RNNs and in Transformers \citep{Liu2016AgreementOT,zhou2019synchronous},
further motivates conditional translation.
This is often explained by lack of context and disregard for the future tokens. Such future context is used by humans \citep{Xia2017DeliberationNS} and can potentially improve NMT \citep{Tu2016ModelingCF, Mi2016CoverageEM}. Moreover, as the encoded input is constant throughout the prediction, the varying performance is likely due to the decoder. Attending to all predictions from lower layers, as we propose here, aims to provide more of this required information.\footnote{Admittedly, for the very first generated tokens, bidirectionality will not help, as there is nothing to attend to.}

Finally, previous work investigated the reasons  why incorporating source syntax helps RNNs \citep{Shi2018OnTN} and Transformers \citep{Pham2019PromotingTK,Sachan2020DoST}. These works show evidence that similar gains can be obtained when incorporating either syntactic trees or non-syntactic, syntactically uninformative, ones.
A hypothesis followed, that graph-like architectures are helpful, but that syntactic information is redundant. While GCN creates such an architecture, linearized syntax, arguably {\sc Parent} and to some extent the labels GCN component, do not. Still, they allow gains over the vanilla decoder, which challenges this hypothesis.

\section{Conclusion}

We presented a flexible method for constructing decoders capable of outputting trees and graphs. We show that the improved decoder achieves notable gains in syntactic generalization, and in some settings improves overall performance as well. 
Our proposal is based on two main modifications to the standard Transformer decoder: (1) autoregression on structure; (2) bidirectional attention in the decoder, which allows recomputing token embeddings in light of newly decoded tokens. Testing on two variants for the decoder, we find that they both show superior syntactic generalization abilities over the vanilla Transformer, and that the gap does not diminish with model size.
The method is flexible enough to allow decoding into a wide variety of graph and tree structures.

Our work opens many avenues for future work. One direction would be to focus on conditional networks, training with (intentionally) noisy prefixes, randomly masking ``predicted'' spans during training \citep[as done in masked language models,][]{devlin-etal-2019-bert}, and data augmentation through hard words or phrases rather than full sentences. 
Another direction might enhance bidirectionality by allowing ``regretting'' and changing past predictions. 
Finally, the work opens possibilities for better incorporating structure into language generators, of incorporating semantic structure and of enforcing meaning preservation \citep[thus targeting hallucinations,][]{wang-sennrich-2020-exposure}, by incorporating source and target structure together.

\section{Acknowledgements}
 We thank Daniel Lehmann which helped in some of the analysis.
The work was done with the support of the Israel Science Foundation (grant no. 929/17) and the Kamin project. 
	
\bibliographystyle{acl_natbib}
\bibliography{references2,references3,references4}

\begin{thebibliography}{82}
\expandafter\ifx\csname natexlab\endcsname\relax\def\natexlab#1{#1}\fi

\bibitem[{Aharoni and Goldberg(2017{\natexlab{a}})}]{aharoni2017morphological}
Roee Aharoni and Yoav Goldberg. 2017{\natexlab{a}}.
\newblock Morphological inflection generation with hard monotonic attention.
\newblock In \emph{Proc. of ACL}, pages 2004--2015.

\bibitem[{Aharoni and Goldberg(2017{\natexlab{b}})}]{Aharoni2017TowardsSN}
Roee Aharoni and Yoav Goldberg. 2017{\natexlab{b}}.
\newblock Towards string-to-tree neural machine translation.
\newblock In \emph{ACL}.

\bibitem[{Avramidis et~al.(2020)Avramidis, Macketanz, Strohriegel, Burchardt,
  and M{\"o}ller}]{avramidis2020fine}
Eleftherios Avramidis, Vivien Macketanz, Ursula Strohriegel, Aljoscha
  Burchardt, and Sebastian M{\"o}ller. 2020.
\newblock Fine-grained linguistic evaluation for state-of-the-art machine
  translation.
\newblock In \emph{Proceedings of the Fifth Conference on Machine Translation},
  pages 346--356.

\bibitem[{Barrault et~al.(2020)Barrault, Biesialska, Bojar, Costa-juss{\`a},
  Federmann, Graham, Grundkiewicz, Haddow, Huck, Joanis, Kocmi, Koehn, kiu Lo,
  Ljubesic, Monz, Morishita, Nagata, Nakazawa, Pal, Post, and
  Zampieri}]{Barrault2020FindingsOT}
Lo{\"i}c Barrault, Magdalena Biesialska, Ondrej Bojar, Marta~R.
  Costa-juss{\`a}, C.~Federmann, Yvette Graham, Roman Grundkiewicz, B.~Haddow,
  Matthias Huck, E.~Joanis, Tom Kocmi, Philipp Koehn, Chi kiu Lo, Nikola
  Ljubesic, Christof Monz, Makoto Morishita, M.~Nagata, T.~Nakazawa, Santanu
  Pal, Matt Post, and Marcos Zampieri. 2020.
\newblock Findings of the 2020 conference on machine translation (wmt20).
\newblock In \emph{WMT}.

\bibitem[{Bastings et~al.(2017)Bastings, Titov, Aziz, Marcheggiani, and
  Sima{\'a}n}]{bastings-etal-2017-graph}
Jasmijn Bastings, Ivan Titov, Wilker Aziz, Diego Marcheggiani, and Khalil
  Sima{\'a}n. 2017.
\newblock Graph convolutional encoders for syntax-aware neural machine
  translation.
\newblock In \emph{Proc. of EMNLP}.

\bibitem[{Belinkov and Bisk(2017)}]{belinkov2018synthetic}
Yonatan Belinkov and Yonatan Bisk. 2017.
\newblock Synthetic and natural noise both break neural machine translation.
\newblock \emph{ICLR}, abs/1711.02173.

\bibitem[{Bisazza et~al.(2021)Bisazza, Ustun, and Sportel}]{Bisazza2021OnTD}
Arianna Bisazza, A.~Ustun, and Stephan Sportel. 2021.
\newblock On the difficulty of translating free-order case-marking languages.
\newblock \emph{ArXiv}, abs/2107.06055.

\bibitem[{Bogin et~al.(2020)Bogin, Subramanian, Gardner, and
  Berant}]{bogin2020latent}
Ben Bogin, Sanjay Subramanian, Matt Gardner, and Jonathan Berant. 2020.
\newblock Latent compositional representations improve systematic
  generalization in grounded question answering.
\newblock \emph{arXiv preprint arXiv:2007.00266}.

\bibitem[{Bojar et~al.(2016)Bojar, Chatterjee, Federmann, Graham, Haddow, Huck,
  Yepes, Koehn, Logacheva, Monz et~al.}]{bojar2016findings}
Ond{\v{r}}ej Bojar, Rajen Chatterjee, Christian Federmann, Yvette Graham, Barry
  Haddow, Matthias Huck, Antonio~Jimeno Yepes, Philipp Koehn, Varvara
  Logacheva, Christof Monz, et~al. 2016.
\newblock Findings of the 2016 conference on machine translation.
\newblock In \emph{Proceedings of the First Conference on Machine Translation:
  Volume 2, Shared Task Papers}, pages 131--198.

\bibitem[{Bugliarello and Okazaki(2020)}]{Bugliarello2020EnhancingMT}
Emanuele Bugliarello and N.~Okazaki. 2020.
\newblock Enhancing machine translation with dependency-aware self-attention.
\newblock In \emph{ACL}.

\bibitem[{Chakraborty et~al.(2018)Chakraborty, Allamanis, and
  Ray}]{Chakraborty2018Tree2TreeNT}
Saikat Chakraborty, Miltiadis Allamanis, and Baishakhi Ray. 2018.
\newblock Tree2tree neural translation model for learning source code changes.
\newblock \emph{ArXiv}, abs/1810.00314.

\bibitem[{Chen et~al.(2017)Chen, Wang, Utiyama, Liu, Tamura, Sumita, and
  Zhao}]{chen-etal-2017-neural}
Kehai Chen, Rui Wang, Masao Utiyama, Lemao Liu, Akihiro Tamura, Eiichiro
  Sumita, and Tiejun Zhao. 2017.
\newblock Neural machine translation with source dependency representation.
\newblock In \emph{Proc. of EMNLP}.

\bibitem[{Chen et~al.(2018)Chen, Wang, Utiyama, Sumita, and
  Zhao}]{chen2018syntax}
Kehai Chen, Rui Wang, Masao Utiyama, Eiichiro Sumita, and Tiejun Zhao. 2018.
\newblock Syntax-directed attention for neural machine translation.
\newblock In \emph{Proc. of AAAI}.

\bibitem[{Choshen and Abend(2019)}]{choshen-abend-2019-automatically}
Leshem Choshen and Omri Abend. 2019.
\newblock \href {https://doi.org/10.18653/v1/K19-1028} {Automatically
  extracting challenge sets for non-local phenomena in neural machine
  translation}.
\newblock In \emph{Proceedings of the 23rd Conference on Computational Natural
  Language Learning (CoNLL)}, pages 291--303, Hong Kong, China. Association for
  Computational Linguistics.

\bibitem[{Choshen et~al.(2020)Choshen, Fox, Aizenbud, and
  Abend}]{Choshen2020OnTW}
Leshem Choshen, Lior Fox, Zohar Aizenbud, and Omri Abend. 2020.
\newblock On the weaknesses of reinforcement learning for neural machine
  translation.
\newblock \emph{ArXiv}, abs/1907.01752.

\bibitem[{Csord{\'a}s et~al.(2021)Csord{\'a}s, Irie, and
  Schmidhuber}]{Csordas2021TheDI}
R{\'o}bert Csord{\'a}s, Kazuki Irie, and J{\"u}rgen Schmidhuber. 2021.
\newblock The devil is in the detail: Simple tricks improve systematic
  generalization of transformers.
\newblock \emph{ArXiv}, abs/2108.12284.

\bibitem[{Devlin et~al.(2019)Devlin, Chang, Lee, and
  Toutanova}]{devlin-etal-2019-bert}
Jacob Devlin, Ming-Wei Chang, Kenton Lee, and Kristina Toutanova. 2019.
\newblock \href {https://doi.org/10.18653/v1/N19-1423} {{BERT}: Pre-training of
  deep bidirectional transformers for language understanding}.
\newblock In \emph{Proceedings of the 2019 Conference of the North {A}merican
  Chapter of the Association for Computational Linguistics: Human Language
  Technologies, Volume 1 (Long and Short Papers)}, pages 4171--4186,
  Minneapolis, Minnesota. Association for Computational Linguistics.

\bibitem[{Ding et~al.(2019)Ding, Renduchintala, and Duh}]{ding2019call}
Shuoyang Ding, Adithya Renduchintala, and Kevin Duh. 2019.
\newblock \href {https://www.aclweb.org/anthology/W19-6620} {A call for prudent
  choice of subword merge operations in neural machine translation}.
\newblock In \emph{Proceedings of Machine Translation Summit XVII Volume 1:
  Research Track}, pages 204--213, Dublin, Ireland. European Association for
  Machine Translation.

\bibitem[{Don-Yehiya et~al.(2022)Don-Yehiya, Choshen, and
  Abend}]{don2022prequel}
Shachar Don-Yehiya, Leshem Choshen, and Omri Abend. 2022.
\newblock Prequel: Quality estimation of machine translation outputs in
  advance.
\newblock \emph{arXiv preprint arXiv:2205.09178}.

\bibitem[{Dyer et~al.(2013)Dyer, Chahuneau, and Smith}]{Dyer2013fastalign}
Chris Dyer, Victor Chahuneau, and Noah~A. Smith. 2013.
\newblock A simple, fast, and effective reparameterization of ibm model 2.
\newblock In \emph{HLT-NAACL}.

\bibitem[{Dyer et~al.(2016)Dyer, Kuncoro, Ballesteros, and
  Smith}]{Dyer2016RecurrentNN}
Chris Dyer, Adhiguna Kuncoro, Miguel Ballesteros, and Noah~A. Smith. 2016.
\newblock Recurrent neural network grammars.
\newblock In \emph{HLT-NAACL}.

\bibitem[{Eriguchi et~al.(2016)Eriguchi, Hashimoto, and
  Tsuruoka}]{eriguchi2016tree}
Akiko Eriguchi, Kazuma Hashimoto, and Yoshimasa Tsuruoka. 2016.
\newblock \href {https://doi.org/10.18653/v1/P16-1078} {Tree-to-sequence
  attentional neural machine translation}.
\newblock In \emph{Proceedings of the 54th Annual Meeting of the Association
  for Computational Linguistics (Volume 1: Long Papers)}, pages 823--833,
  Berlin, Germany. Association for Computational Linguistics.

\bibitem[{Eriguchi et~al.(2017)Eriguchi, Tsuruoka, and
  Cho}]{eriguchi2017learningTP}
Akiko Eriguchi, Yoshimasa Tsuruoka, and Kyunghyun Cho. 2017.
\newblock Learning to parse and translate improves neural machine translation.
\newblock \emph{ArXiv}, abs/1702.03525.

\bibitem[{Fern{\'a}ndez-Gonz{\'a}lez and
  G{\'o}mez-Rodr{\'\i}guez(2018)}]{fernandez2018nonprojective}
Daniel Fern{\'a}ndez-Gonz{\'a}lez and Carlos G{\'o}mez-Rodr{\'\i}guez. 2018.
\newblock \href {https://doi.org/10.18653/v1/N18-2109} {Non-projective
  dependency parsing with non-local transitions}.
\newblock In \emph{Proceedings of the 2018 Conference of the North {A}merican
  Chapter of the Association for Computational Linguistics: Human Language
  Technologies, Volume 2 (Short Papers)}, pages 693--700, New Orleans,
  Louisiana. Association for Computational Linguistics.

\bibitem[{Gontier et~al.(2020)Gontier, Sinha, Reddy, and
  Pal}]{gontier2020measuring}
Nicolas Gontier, Koustuv Sinha, Siva Reddy, and Christopher Pal. 2020.
\newblock Measuring systematic generalization in neural proof generation with
  transformers.
\newblock \emph{arXiv preprint arXiv:2009.14786}.

\bibitem[{G{\=u} et~al.(2018)G{\=u}, Shavarani, and Sarkar}]{gu-etal-2018-top}
Jetic G{\=u}, Hassan~S. Shavarani, and Anoop Sarkar. 2018.
\newblock \href {https://doi.org/10.18653/v1/D18-1037} {Top-down tree
  structured decoding with syntactic connections for neural machine translation
  and parsing}.
\newblock In \emph{Proceedings of the 2018 Conference on Empirical Methods in
  Natural Language Processing}, pages 401--413, Brussels, Belgium. Association
  for Computational Linguistics.

\bibitem[{Hao et~al.(2019)Hao, Wang, Shi, Zhang, and Tu}]{hao2019multi}
Jie Hao, Xing Wang, Shuming Shi, Jinfeng Zhang, and Zhaopeng Tu. 2019.
\newblock \href {https://doi.org/10.18653/v1/D19-1082} {Multi-granularity
  self-attention for neural machine translation}.
\newblock In \emph{Proceedings of the 2019 Conference on Empirical Methods in
  Natural Language Processing and the 9th International Joint Conference on
  Natural Language Processing (EMNLP-IJCNLP)}, pages 887--897, Hong Kong,
  China. Association for Computational Linguistics.

\bibitem[{Harer et~al.(2019)Harer, Reale, and
  Chin}]{Harer2019TreeTransformerAT}
Jacob Harer, C.~Reale, and P.~Chin. 2019.
\newblock Tree-transformer: A transformer-based method for correction of
  tree-structured data.
\newblock \emph{ArXiv}, abs/1908.00449.

\bibitem[{Hershcovich et~al.(2018)Hershcovich, Abend, and
  Rappoport}]{hershcovich2018multitask}
Daniel Hershcovich, Omri Abend, and Ari Rappoport. 2018.
\newblock Multitask parsing across semantic representations.
\newblock In \emph{Proc. of ACL}, pages 373--385.

\bibitem[{Hu et~al.(2020)Hu, Gauthier, Qian, Wilcox, and
  Levy}]{hu2020systematic}
Jennifer Hu, Jon Gauthier, Peng Qian, Ethan Wilcox, and Roger Levy. 2020.
\newblock A systematic assessment of syntactic generalization in neural
  language models.
\newblock In \emph{Proceedings of the 58th Annual Meeting of the Association
  for Computational Linguistics}, pages 1725--1744.

\bibitem[{Jawahar et~al.(2019)Jawahar, Sagot, and Seddah}]{jawahar2019bert}
Ganesh Jawahar, Beno{\^\i}t Sagot, and Djam{\'e} Seddah. 2019.
\newblock \href {https://doi.org/10.18653/v1/P19-1356} {What does {BERT} learn
  about the structure of language?}
\newblock In \emph{Proceedings of the 57th Annual Meeting of the Association
  for Computational Linguistics}, pages 3651--3657, Florence, Italy.
  Association for Computational Linguistics.

\bibitem[{Kingma and Ba(2015)}]{Kingma2015AdamAM}
Diederik~P. Kingma and Jimmy Ba. 2015.
\newblock Adam: A method for stochastic optimization.
\newblock \emph{CoRR}, abs/1412.6980.

\bibitem[{Kipf and Welling(2016)}]{kipf2016semi}
Thomas~N. Kipf and Max Welling. 2016.
\newblock \href {http://arxiv.org/abs/1609.02907} {Semi-supervised
  classification with graph convolutional networks}.
\newblock \emph{CoRR}, abs/1609.02907.

\bibitem[{Koehn et~al.(2007)Koehn, Hoang, Birch, Callison-Burch, Federico,
  Bertoldi, Cowan, Shen, Moran, Zens, Dyer, Bojar, Constantin, and
  Herbst}]{Koehn2007Moses}
Philipp Koehn, Hieu Hoang, Alexandra Birch, Chris Callison-Burch, M.~Federico,
  N.~Bertoldi, B.~Cowan, Wade Shen, C.~Moran, R.~Zens, Chris Dyer, Ondrej
  Bojar, A.~Constantin, and E.~Herbst. 2007.
\newblock Moses: Open source toolkit for statistical machine translation.
\newblock In \emph{ACL}.

\bibitem[{Li et~al.(2017)Li, Xiong, Tu, Zhu, Zhang, and
  Zhou}]{li-etal-2017-modeling}
Junhui Li, Deyi Xiong, Zhaopeng Tu, Muhua Zhu, Min Zhang, and Guodong Zhou.
  2017.
\newblock Modeling source syntax for neural machine translation.
\newblock In \emph{Proc. of ACL}.

\bibitem[{Linzen and Baroni(2020)}]{linzen2020syntactic}
Tal Linzen and Marco Baroni. 2020.
\newblock Syntactic structure from deep learning.
\newblock \emph{Annual Review of Linguistics}, 7.

\bibitem[{Liu et~al.(2016)Liu, Utiyama, Finch, and Sumita}]{Liu2016AgreementOT}
L.~Liu, M.~Utiyama, A.~Finch, and Eiichiro Sumita. 2016.
\newblock Agreement on target-bidirectional neural machine translation.
\newblock In \emph{HLT-NAACL}.

\bibitem[{Lopez(2008)}]{Lopez:08}
Adam Lopez. 2008.
\newblock Statistical machine translation.
\newblock \emph{ACM Computing Surveys (CSUR)}, 40:8.

\bibitem[{Lui and Baldwin(2012)}]{lui2012langid}
Marco Lui and Timothy Baldwin. 2012.
\newblock langid. py: An off-the-shelf language identification tool.
\newblock In \emph{Proceedings of the ACL 2012 system demonstrations}, pages
  25--30.

\bibitem[{Ma et~al.(2019)Ma, Wei, Bojar, and Graham}]{ma2019wmt19metrics}
Qingsong Ma, Johnny Wei, Ond{\v{r}}ej Bojar, and Yvette Graham. 2019.
\newblock \href {https://doi.org/10.18653/v1/W19-5302} {Results of the {WMT}19
  metrics shared task: Segment-level and strong {MT} systems pose big
  challenges}.
\newblock In \emph{Proceedings of the Fourth Conference on Machine Translation
  (Volume 2: Shared Task Papers, Day 1)}, pages 62--90, Florence, Italy.
  Association for Computational Linguistics.

\bibitem[{Manning et~al.(2020)Manning, Clark, Hewitt, Khandelwal, and
  Levy}]{manning2020emergent}
Christopher~D. Manning, Kevin Clark, John Hewitt, Urvashi Khandelwal, and Omer
  Levy. 2020.
\newblock \href {https://doi.org/10.1073/pnas.1907367117} {Emergent linguistic
  structure in artificial neural networks trained by self-supervision}.
\newblock \emph{PNAS}.

\bibitem[{Marcheggiani et~al.(2018)Marcheggiani, Bastings, and
  Titov}]{marcheggiani-etal-2018-exploiting}
Diego Marcheggiani, Jasmijn Bastings, and Ivan Titov. 2018.
\newblock Exploiting semantics in neural machine translation with graph
  convolutional networks.
\newblock In \emph{Proc. of NAACL}.

\bibitem[{McCoy et~al.(2019)McCoy, Pavlick, and Linzen}]{mccoy2019right}
Tom McCoy, Ellie Pavlick, and Tal Linzen. 2019.
\newblock Right for the wrong reasons: Diagnosing syntactic heuristics in
  natural language inference.
\newblock In \emph{Proceedings of the 57th Annual Meeting of the Association
  for Computational Linguistics}, pages 3428--3448.

\bibitem[{Mi et~al.(2016)Mi, Sankaran, Wang, and
  Ittycheriah}]{Mi2016CoverageEM}
Haitao Mi, B.~Sankaran, Z.~Wang, and Abe Ittycheriah. 2016.
\newblock Coverage embedding models for neural machine translation.
\newblock In \emph{EMNLP}.

\bibitem[{Nadejde et~al.(2017)Nadejde, Reddy, Sennrich, Dwojak,
  Junczys-Dowmunt, Koehn, and Birch}]{Nadejde2017PredictingTL}
Maria Nadejde, Siva Reddy, Rico Sennrich, Tomasz Dwojak, Marcin
  Junczys-Dowmunt, P.~Koehn, and Alexandra Birch. 2017.
\newblock Predicting target language ccg supertags improves neural machine
  translation.
\newblock In \emph{WMT}.

\bibitem[{Nguyen et~al.(2020)Nguyen, Joty, Hoi, and
  Socher}]{Nguyen2020TreestructuredAW}
Xuan-Phi Nguyen, Shafiq~R. Joty, S.~Hoi, and R.~Socher. 2020.
\newblock Tree-structured attention with hierarchical accumulation.
\newblock \emph{ArXiv}, abs/2002.08046.

\bibitem[{Nivre(2003)}]{Nivre2003AnEA}
Joakim Nivre. 2003.
\newblock An efficient algorithm for projective dependency parsing.
\newblock In \emph{IWPT}.

\bibitem[{Nivre et~al.(2016)Nivre, de~Marneffe, Ginter, Goldberg, Hajic,
  Manning, McDonald, Petrov, Pyysalo, Silveira, Tsarfaty, and
  Zeman}]{nivre2016universal}
Joakim Nivre, Marie-Catherine de~Marneffe, Filip Ginter, Yoav Goldberg, Jan
  Hajic, Christopher~D. Manning, Ryan McDonald, Slav Petrov, Sampo Pyysalo,
  Natalia Silveira, Reut Tsarfaty, and Daniel Zeman. 2016.
\newblock Universal dependencies v1: A multilingual treebank collection.
\newblock In \emph{Proc. of LREC}, pages 1659--1666.

\bibitem[{Oepen et~al.(2020)Oepen, Abend, Abzianidze, Bos, Hajic, Hershcovich,
  Li, O{'}Gorman, Xue, and Zeman}]{oepen-etal-2020-mrp}
Stephan Oepen, Omri Abend, Lasha Abzianidze, Johan Bos, Jan Hajic, Daniel
  Hershcovich, Bin Li, Tim O{'}Gorman, Nianwen Xue, and Daniel Zeman. 2020.
\newblock \href {https://doi.org/10.18653/v1/2020.conll-shared.1} {{MRP} 2020:
  The second shared task on cross-framework and cross-lingual meaning
  representation parsing}.
\newblock In \emph{Proceedings of the CoNLL 2020 Shared Task: Cross-Framework
  Meaning Representation Parsing}, pages 1--22, Online. Association for
  Computational Linguistics.

\bibitem[{Ontan{\'o}n et~al.(2021)Ontan{\'o}n, Ainslie, Cvicek, and
  Fisher}]{Ontanon2021MakingTS}
Santiago Ontan{\'o}n, Joshua Ainslie, V.~Cvicek, and Zachary~Kenneth Fisher.
  2021.
\newblock Making transformers solve compositional tasks.
\newblock \emph{ArXiv}, abs/2108.04378.

\bibitem[{Papineni et~al.(2002)Papineni, Roukos, Ward, and
  Zhu}]{Papineni2002BleuAM}
Kishore Papineni, S.~Roukos, T.~Ward, and Wei-Jing Zhu. 2002.
\newblock Bleu: a method for automatic evaluation of machine translation.
\newblock In \emph{ACL}.

\bibitem[{Pham et~al.(2019)Pham, Mach{\'a}cek, and Bojar}]{Pham2019PromotingTK}
Thuong-Hai Pham, Dominik Mach{\'a}cek, and Ondrej Bojar. 2019.
\newblock Promoting the knowledge of source syntax in transformer nmt is not
  needed.
\newblock \emph{Computaci{\'o}n y Sistemas}, 23.

\bibitem[{Ponti et~al.(2018)Ponti, Reichart, Korhonen, and
  Vuli{\'c}}]{ponti2018isomorphic}
Edoardo~Maria Ponti, Roi Reichart, Anna Korhonen, and Ivan Vuli{\'c}. 2018.
\newblock Isomorphic transfer of syntactic structures in cross-lingual nlp.
\newblock In \emph{Proc. of ACL}, volume~1.

\bibitem[{Popel and Bojar(2018)}]{popel2018training}
Martin Popel and Ond{\v{r}}ej Bojar. 2018.
\newblock Training tips for the transformer model.
\newblock \emph{The Prague Bulletin of Mathematical Linguistics},
  110(1):43--70.

\bibitem[{Popovic(2017)}]{Popovic2017chrFWH}
Maja Popovic. 2017.
\newblock chrf++: words helping character n-grams.
\newblock In \emph{WMT}.

\bibitem[{Ravfogel et~al.(2019)Ravfogel, Goldberg, and
  Linzen}]{Ravfogel2019StudyingTI}
Shauli Ravfogel, Y.~Goldberg, and Tal Linzen. 2019.
\newblock Studying the inductive biases of rnns with synthetic variations of
  natural languages.
\newblock \emph{ArXiv}, abs/1903.06400.

\bibitem[{Sachan et~al.(2020)Sachan, Zhang, Qi, and Hamilton}]{Sachan2020DoST}
D.~Sachan, Yuhao Zhang, Peng Qi, and W.~Hamilton. 2020.
\newblock Do syntax trees help pre-trained transformers extract information?
\newblock \emph{ArXiv}, abs/2008.09084.

\bibitem[{Sennrich et~al.(2017)Sennrich, Firat, Cho, Birch, Haddow, Hitschler,
  Junczys-Dowmunt, L{\"a}ubli, Barone, Mokry, and
  Nadejde}]{Sennrich2017NematusAT}
Rico Sennrich, Orhan Firat, K.~Cho, Alexandra Birch, B.~Haddow, Julian
  Hitschler, Marcin Junczys-Dowmunt, Samuel L{\"a}ubli, A.~Barone, Jozef Mokry,
  and Maria Nadejde. 2017.
\newblock Nematus: a toolkit for neural machine translation.
\newblock In \emph{EACL}.

\bibitem[{Sennrich et~al.(2016)Sennrich, Haddow, and
  Birch}]{sennrich2016neural}
Rico Sennrich, Barry Haddow, and Alexandra Birch. 2016.
\newblock \href {https://doi.org/10.18653/v1/P16-1162} {Neural machine
  translation of rare words with subword units}.
\newblock In \emph{Proceedings of the 54th Annual Meeting of the Association
  for Computational Linguistics (Volume 1: Long Papers)}, pages 1715--1725,
  Berlin, Germany. Association for Computational Linguistics.

\bibitem[{Shi et~al.(2018)Shi, Zhou, Chen, and Li}]{Shi2018OnTN}
Haoyue Shi, Hao Zhou, J.~Chen, and Lei Li. 2018.
\newblock On tree-based neural sentence modeling.
\newblock In \emph{EMNLP}.

\bibitem[{Shiv and Quirk(2019)}]{Shiv2019NovelPE}
Vighnesh~Leonardo Shiv and Chris Quirk. 2019.
\newblock Novel positional encodings to enable tree-based transformers.
\newblock In \emph{NeurIPS}.

\bibitem[{Song et~al.(2019)Song, Gildea, Zhang, Wang, and
  Su}]{song-etal-2019-semantic}
Linfeng Song, Daniel Gildea, Yue Zhang, Zhiguo Wang, and Jinsong Su. 2019.
\newblock Semantic neural machine translation using {AMR}.
\newblock \emph{TACL}, 7.

\bibitem[{Stanojevi{\'c} and Steedman(2020)}]{stanojevic-steedman-2020-max}
Milo{\v{s}} Stanojevi{\'c} and Mark Steedman. 2020.
\newblock \href {https://doi.org/10.18653/v1/2020.acl-main.378} {Max-margin
  incremental {CCG} parsing}.
\newblock In \emph{Proceedings of the 58th Annual Meeting of the Association
  for Computational Linguistics}, pages 4111--4122, Online. Association for
  Computational Linguistics.

\bibitem[{Straka(2018)}]{straka2018udpipe}
Milan Straka. 2018.
\newblock Udpipe 2.0 prototype at conll 2018 ud shared task.
\newblock In \emph{Proceedings of the CoNLL 2018 Shared Task: Multilingual
  Parsing from Raw Text to Universal Dependencies}, pages 197--207.

\bibitem[{Strubell et~al.(2018)Strubell, Verga, Andor, Weiss, and
  McCallum}]{strubell-etal-2018-linguistically}
Emma Strubell, Patrick Verga, Daniel Andor, David Weiss, and Andrew McCallum.
  2018.
\newblock \href {https://doi.org/10.18653/v1/D18-1548} {Linguistically-informed
  self-attention for semantic role labeling}.
\newblock In \emph{Proceedings of the 2018 Conference on Empirical Methods in
  Natural Language Processing}, pages 5027--5038, Brussels, Belgium.
  Association for Computational Linguistics.

\bibitem[{Tai et~al.(2015)Tai, Socher, and Manning}]{Tai2015ImprovedSR}
Kai~Sheng Tai, R.~Socher, and Christopher~D. Manning. 2015.
\newblock Improved semantic representations from tree-structured long
  short-term memory networks.
\newblock In \emph{ACL}.

\bibitem[{Tiedemann(2012)}]{Tiedemann2012opus}
J.~Tiedemann. 2012.
\newblock Parallel data, tools and interfaces in opus.
\newblock In \emph{LREC}.

\bibitem[{Tu et~al.(2016)Tu, Lu, Liu, Liu, and Li}]{Tu2016ModelingCF}
Zhaopeng Tu, Z.~Lu, Y.~Liu, X.~Liu, and Hang Li. 2016.
\newblock Modeling coverage for neural machine translation.
\newblock \emph{arXiv: Computation and Language}.

\bibitem[{Vaswani et~al.(2017)Vaswani, Shazeer, Parmar, Uszkoreit, Jones,
  Gomez, Kaiser, and Polosukhin}]{vaswani2017attention}
Ashish Vaswani, Noam Shazeer, Niki Parmar, Jakob Uszkoreit, Llion Jones,
  Aidan~N Gomez, {\L}ukasz Kaiser, and Illia Polosukhin. 2017.
\newblock \href
  {https://papers.nips.cc/paper/7181-attention-is-all-you-need.pdf} {Attention
  is all you need}.
\newblock In \emph{Advances in Neural Information Processing Systems}, pages
  5998--6008.

\bibitem[{Wang and Sennrich(2020)}]{wang-sennrich-2020-exposure}
Chaojun Wang and Rico Sennrich. 2020.
\newblock \href {https://doi.org/10.18653/v1/2020.acl-main.326} {On exposure
  bias, hallucination and domain shift in neural machine translation}.
\newblock In \emph{Proceedings of the 58th Annual Meeting of the Association
  for Computational Linguistics}, pages 3544--3552, Online. Association for
  Computational Linguistics.

\bibitem[{Wang et~al.(2018)Wang, Pham, Yin, and Neubig}]{wang2018tree}
Xinyi Wang, Hieu Pham, Pengcheng Yin, and Graham Neubig. 2018.
\newblock A tree-based decoder for neural machine translation.
\newblock \emph{arXiv preprint arXiv:1808.09374}.

\bibitem[{Wang et~al.(2019)Wang, yi~Lee, and Chen}]{Wang2019TreeTI}
Yau-Shian Wang, Hung yi~Lee, and Yun-Nung Chen. 2019.
\newblock Tree transformer: Integrating tree structures into self-attention.
\newblock In \emph{EMNLP/IJCNLP}.

\bibitem[{Welleck et~al.(2021)Welleck, West, Cao, and
  Choi}]{welleck2021symbolic}
Sean Welleck, Peter West, Jize Cao, and Yejin Choi. 2021.
\newblock Symbolic brittleness in sequence models: on systematic generalization
  in symbolic mathematics.
\newblock \emph{arXiv preprint arXiv:2109.13986}.

\bibitem[{Wieting et~al.(2019)Wieting, Berg-Kirkpatrick, Gimpel, and
  Neubig}]{Wieting2019BeyondBT}
J.~Wieting, Taylor Berg-Kirkpatrick, Kevin Gimpel, and Graham Neubig. 2019.
\newblock Beyond bleu: Training neural machine translation with semantic
  similarity.
\newblock In \emph{ACL}.

\bibitem[{Wilcox et~al.(2020)Wilcox, Qian, Futrell, Kohita, Levy, and
  Ballesteros}]{wilcox2020structural}
Ethan Wilcox, Peng Qian, Richard Futrell, Ryosuke Kohita, Roger Levy, and
  Miguel Ballesteros. 2020.
\newblock Structural supervision improves few-shot learning and syntactic
  generalization in neural language models.
\newblock \emph{arXiv preprint arXiv:2010.05725}.

\bibitem[{Wu et~al.(2017)Wu, Zhang, Yang, Li, and Zhou}]{wu-etal-2017-sequence}
Shuangzhi Wu, Dongdong Zhang, Nan Yang, Mu~Li, and Ming Zhou. 2017.
\newblock Sequence-to-dependency neural machine translation.
\newblock In \emph{Proc. of ACL}.

\bibitem[{Xia et~al.(2017)Xia, Tian, Wu, Lin, Qin, Yu, and
  Liu}]{Xia2017DeliberationNS}
Yingce Xia, Fei Tian, Lijun Wu, Jianxin Lin, T.~Qin, N.~Yu, and T.~Liu. 2017.
\newblock Deliberation networks: Sequence generation beyond one-pass decoding.
\newblock In \emph{NIPS}.

\bibitem[{Yehudai et~al.(2022)Yehudai, Choshen, Fox, and Abend}]{yehudai}
Asaf Yehudai, Leshem Choshen, Lior Fox, and Omri Abend. 2022.
\newblock Reinforcement learning with large action spaces for neural machine
  translation.
\newblock In \emph{COLING}.

\bibitem[{Zhang et~al.(2019)Zhang, Li, Fu, and Zhang}]{zhang2019syntaxEnhanced}
Meishan Zhang, Zhenghua Li, Guohong Fu, and Min Zhang. 2019.
\newblock \href {https://doi.org/10.18653/v1/N19-1118} {Syntax-enhanced neural
  machine translation with syntax-aware word representations}.
\newblock In \emph{Proceedings of the 2019 Conference of the North {A}merican
  Chapter of the Association for Computational Linguistics: Human Language
  Technologies, Volume 1 (Long and Short Papers)}, pages 1151--1161,
  Minneapolis, Minnesota. Association for Computational Linguistics.

\bibitem[{Zhang et~al.(2018)Zhang, Su, Qin, Liu, Ji, and
  Wang}]{Zhang2018AsynchronousBD}
Xiangwen Zhang, Jinsong Su, Yue Qin, Y.~Liu, R.~Ji, and Hongji Wang. 2018.
\newblock Asynchronous bidirectional decoding for neural machine translation.
\newblock \emph{ArXiv}, abs/1801.05122.

\bibitem[{Zhou et~al.(2019{\natexlab{a}})Zhou, Ma, Hu, and
  Neubig}]{zhou-etal-2019-handling}
Chunting Zhou, Xuezhe Ma, Junjie Hu, and Graham Neubig. 2019{\natexlab{a}}.
\newblock Handling syntactic divergence in low-resource machine translation.
\newblock In \emph{Proc. of EMNLP-IJCNLP}, pages 1388--1394.

\bibitem[{Zhou et~al.(2019{\natexlab{b}})Zhou, Zhang, and
  Zong}]{zhou2019synchronous}
Long Zhou, Jiajun Zhang, and Chengqing Zong. 2019{\natexlab{b}}.
\newblock Synchronous bidirectional neural machine translation.
\newblock \emph{Transactions of the Association for Computational Linguistics},
  7:91--105.

\end{thebibliography}


\begin{thebibliography}{15}
\expandafter\ifx\csname natexlab\endcsname\relax\def\natexlab#1{#1}\fi

\bibitem[{Bisazza et~al.(2021)Bisazza, Ustun, and Sportel}]{Bisazza2021OnTD}
Arianna Bisazza, A.~Ustun, and Stephan Sportel. 2021.
\newblock On the difficulty of translating free-order case-marking languages.
\newblock \emph{ArXiv}, abs/2107.06055.

\bibitem[{Ding et~al.(2019)Ding, Renduchintala, and Duh}]{ding2019call}
Shuoyang Ding, Adithya Renduchintala, and Kevin Duh. 2019.
\newblock \href {https://www.aclweb.org/anthology/W19-6620} {A call for prudent
  choice of subword merge operations in neural machine translation}.
\newblock In \emph{Proceedings of Machine Translation Summit XVII Volume 1:
  Research Track}, pages 204--213, Dublin, Ireland. European Association for
  Machine Translation.

\bibitem[{Dyer et~al.(2013)Dyer, Chahuneau, and Smith}]{Dyer2013fastalign}
Chris Dyer, Victor Chahuneau, and Noah~A. Smith. 2013.
\newblock A simple, fast, and effective reparameterization of ibm model 2.
\newblock In \emph{HLT-NAACL}.

\bibitem[{Fern{\'a}ndez-Gonz{\'a}lez and
  G{\'o}mez-Rodr{\'\i}guez(2018)}]{fernandez2018nonprojective}
Daniel Fern{\'a}ndez-Gonz{\'a}lez and Carlos G{\'o}mez-Rodr{\'\i}guez. 2018.
\newblock \href {https://doi.org/10.18653/v1/N18-2109} {Non-projective
  dependency parsing with non-local transitions}.
\newblock In \emph{Proceedings of the 2018 Conference of the North {A}merican
  Chapter of the Association for Computational Linguistics: Human Language
  Technologies, Volume 2 (Short Papers)}, pages 693--700, New Orleans,
  Louisiana. Association for Computational Linguistics.

\bibitem[{Kingma and Ba(2015)}]{Kingma2015AdamAM}
Diederik~P. Kingma and Jimmy Ba. 2015.
\newblock Adam: A method for stochastic optimization.
\newblock \emph{CoRR}, abs/1412.6980.

\bibitem[{Koehn et~al.(2007)Koehn, Hoang, Birch, Callison-Burch, Federico,
  Bertoldi, Cowan, Shen, Moran, Zens, Dyer, Bojar, Constantin, and
  Herbst}]{Koehn2007Moses}
Philipp Koehn, Hieu Hoang, Alexandra Birch, Chris Callison-Burch, M.~Federico,
  N.~Bertoldi, B.~Cowan, Wade Shen, C.~Moran, R.~Zens, Chris Dyer, Ondrej
  Bojar, A.~Constantin, and E.~Herbst. 2007.
\newblock Moses: Open source toolkit for statistical machine translation.
\newblock In \emph{ACL}.

\bibitem[{Lui and Baldwin(2012)}]{lui2012langid}
Marco Lui and Timothy Baldwin. 2012.
\newblock langid. py: An off-the-shelf language identification tool.
\newblock In \emph{Proceedings of the ACL 2012 system demonstrations}, pages
  25--30.

\bibitem[{Ma et~al.(2019)Ma, Wei, Bojar, and Graham}]{ma2019wmt19metrics}
Qingsong Ma, Johnny Wei, Ond{\v{r}}ej Bojar, and Yvette Graham. 2019.
\newblock \href {https://doi.org/10.18653/v1/W19-5302} {Results of the {WMT}19
  metrics shared task: Segment-level and strong {MT} systems pose big
  challenges}.
\newblock In \emph{Proceedings of the Fourth Conference on Machine Translation
  (Volume 2: Shared Task Papers, Day 1)}, pages 62--90, Florence, Italy.
  Association for Computational Linguistics.

\bibitem[{Papineni et~al.(2002)Papineni, Roukos, Ward, and
  Zhu}]{Papineni2002BleuAM}
Kishore Papineni, S.~Roukos, T.~Ward, and Wei-Jing Zhu. 2002.
\newblock Bleu: a method for automatic evaluation of machine translation.
\newblock In \emph{ACL}.

\bibitem[{Popel and Bojar(2018)}]{popel2018training}
Martin Popel and Ond{\v{r}}ej Bojar. 2018.
\newblock Training tips for the transformer model.
\newblock \emph{The Prague Bulletin of Mathematical Linguistics},
  110(1):43--70.

\bibitem[{Popovic(2017)}]{Popovic2017chrFWH}
Maja Popovic. 2017.
\newblock chrf++: words helping character n-grams.
\newblock In \emph{WMT}.

\bibitem[{Sennrich et~al.(2017)Sennrich, Firat, Cho, Birch, Haddow, Hitschler,
  Junczys-Dowmunt, L{\"a}ubli, Barone, Mokry, and
  Nadejde}]{Sennrich2017NematusAT}
Rico Sennrich, Orhan Firat, K.~Cho, Alexandra Birch, B.~Haddow, Julian
  Hitschler, Marcin Junczys-Dowmunt, Samuel L{\"a}ubli, A.~Barone, Jozef Mokry,
  and Maria Nadejde. 2017.
\newblock Nematus: a toolkit for neural machine translation.
\newblock In \emph{EACL}.

\bibitem[{Sennrich et~al.(2016)Sennrich, Haddow, and
  Birch}]{sennrich2016neural}
Rico Sennrich, Barry Haddow, and Alexandra Birch. 2016.
\newblock \href {https://doi.org/10.18653/v1/P16-1162} {Neural machine
  translation of rare words with subword units}.
\newblock In \emph{Proceedings of the 54th Annual Meeting of the Association
  for Computational Linguistics (Volume 1: Long Papers)}, pages 1715--1725,
  Berlin, Germany. Association for Computational Linguistics.

\bibitem[{Stanojevi{\'c} and Steedman(2020)}]{stanojevic-steedman-2020-max}
Milo{\v{s}} Stanojevi{\'c} and Mark Steedman. 2020.
\newblock \href {https://doi.org/10.18653/v1/2020.acl-main.378} {Max-margin
  incremental {CCG} parsing}.
\newblock In \emph{Proceedings of the 58th Annual Meeting of the Association
  for Computational Linguistics}, pages 4111--4122, Online. Association for
  Computational Linguistics.

\bibitem[{Zhang et~al.(2018)Zhang, Su, Qin, Liu, Ji, and
  Wang}]{Zhang2018AsynchronousBD}
Xiangwen Zhang, Jinsong Su, Yue Qin, Y.~Liu, R.~Ji, and Hongji Wang. 2018.
\newblock Asynchronous bidirectional decoding for neural machine translation.
\newblock \emph{ArXiv}, abs/1801.05122.

\end{thebibliography}

\clearpage

\appendix

\section{From sequence-to-sequence to conditional}\label{sec:conditional}
Attention-based models are characterized by being state-less. They can, therefore, be viewed as conditional language models, namely as models for producing a distribution for the next word, given the generated prefix and source sentence
It is possible to re-encode other information (not only the decoded output) into the decoder at each step, or predict only tokens of interest, rather than the complete sequence.
It is also possible to change the source sentence partially or completely (e.g., adding noise to increase robustness), condition on additional information (\S\ref{sec:synth-arch}) and adjust this information during prediction (e.g. force predicted word characteristics). 
Nevertheless, the standard practice is to only re-encode past predictions.\footnote{This is true even in cases of bidirectional generation \citep[e.g.,][]{Zhang2018AsynchronousBD}.}

Unlike RNNs, attention-based models do not inherently rely on past predictions in terms of inputs, weights and gradients. The only connection to past predictions is mediated through their re-encoding back into the decoder. 

RNNs receive past states as inputs. Backpropagation through time sees the current network as connected to the previous networks supplying the state input. Thus, the gradients take into account past predictions as well.

In contrast, Transformers have gradients over representation of past words only if they are fed into the network. Unlike backpropagation through time, the preceding tokens can be changed, or even omitted (e.g., in a limited window size scenario). Specifically, in our case, preceding tokens may have different representations at each generation step.

To sum, the representation is updated to provide good representation for the current step, but it is not calculated over the actual network of the previous step. It is often the case, though, that the previous decoded words are auto-regressed and hence updated. 

This architecture, therefore, allows more flexibility than RNNs. Still, Transformers are often thought about as an extension to RNNs, i.e., sequnce-to-sequence models. For that reason it is rare to find changes to the training schedule that incorporate more knowledge, change "past" information or translate only parts of a sentence with a network.
With such methods, for example, one can dynamically force features of the next prediction (by a changing input) or augment learning by teaching the network only over hard cases. Such an approach may choose augmented data in a regular way, but stop the prediction at the part in the sentence one wishes the network to learn, or even teach it several alternatives with the same prefix.

\section{Experimental Setup}\label{sec:setup}
The code is adapted from the NEMATUS code repository \citep{Sennrich2017NematusAT} and will be released upon publication. All hyperparameters are either taken from the original suggestions or optimized for the vanilla Transformer and used as is for our suggested models.

Networks are all trained with batch size 128, embedding size 256, 4 decoder and encoder blocks, 8 attention heads (one of which might be a parent head \S\ref{sec:parent}), 90K steps (where empirically some saturation is reached. This is a relatively fair comparison \citep{popel2018training}), learning rate $1e^{-4}$, 4K warm-up steps, Adam \citep{Kingma2015AdamAM} optimizer with beta 0.9 and 0.999 for the first and second moment and epsilon of $1e^{-8}$. We use the standard (structure-unaware) Transformer encoder in all our experiments. 
Each model was trained on 4 NVIDIA Tesla M60 or RTX 2080Ti GPUs for approximately a week (2 for GCN architecture), large models on RTX6000.

Preprocessing includes truecasing, tokenization as implemented by Moses \citep{Koehn2007Moses} and byte pair encoding \citep{sennrich2016neural} without tying. Empty source or target sentences were dropped. In training, the maximum target sentence length is 40 non-transition tokens (BPE).

We used UDPipe English and German over UD 2.0 and Russian with 2.5 with syntagrus version.

In unreported trials, we found that whenever noisy and crawled data is used, filtering is crucial for even the baselines to show reasonable results. On full En-Ru (See \S\ref{sec:overall}), we filter unexpected languages by langID \citep{lui2012langid} and improbable alignment ($p<-180$) with FastAlign \citep{Dyer2013fastalign}. Overall, about half the sentences were filtered by those measures or length. 

There were 4,066,323 training sentences after filtering En-De and 4,468,840 before. In En-Ru, there were 19,557,568 after and 37,948,456 before. The English challenge sets on books and news sizes are respectively, 1,188 and 11 reflexive, 3,953 and 17 particle, 191 and 8 prepositions stranding, and the German 2,628 and 261 reflexive and 7,584 and 232 particle. WMT dev and test sets are always of about 3K sentences in size.

We use {\it chrF++.py} with 1 word and beta of 3 to obtain chrF+ \citep{Popovic2017chrFWH} score as in WMT19 \citep{ma2019wmt19metrics} and detokenized BLEU \citep{Papineni2002BleuAM} as implemented in Moses. We use two automatic metrics: BLEU as the standard measure and chrF+ as it was shown to better correlate with human judgments, while still being simple and understandable \citep{ma2019wmt19metrics}. Both metrics rely on n-gram overlap between the source and reference, where BLEU focuses on word precision, and chrF+ balances precision and recall and includes characters, as well as word n-grams.

\paragraph{Transitions.}
\cready{maybe add parts from here back to the paper}
We made two practical choices when creating the transition graph. First, we deleted the root edge, as the root is not a word in the translation. Second, we train only on projective parses. This choice reduces noise due to the low reliability of current non-projective parsers \citep{fernandez2018nonprojective}, while not losing many training sentences. We do note, however, that this choice is not without its risks: it might be less fitting for some languages in which non-projective sentences are common.

The transitions serve as the NMT vocabulary. There are 45 labels and two directions of connections, summing up to 90 new tokens.
This hardly affects the standard vocabulary size, which usually consists of tens of thousands of tokens\citep{ding2019call}. \cready{cite assaf's paper?}
We treat both token and transition predictions in the same way, and do not rescale their score as done in \citet{stanojevic-steedman-2020-max}.
If anything, the need to memorize more should hurt performance, and so increased performance should come despite enlarging the vocabulary and not because of it. 
It is possible to split the tokens into directions and labels (summing to 47). This comes at the cost of lengthy sentences which increase training time and memory consumption. We did not experiment with other methods for encoding the transitions (e.g., embedding labels and edges separately).

\section{Mixup challenge} \label{sec:mixup}
We follow the results of \citep{Bisazza2021OnTD} that Transformers are able to learn languages with free order, given case markings. Given those findings, we wonder whether indeed Transformers are robust to mixing the order where case marking exists.

To do that, we take lists of nouns and verbs to create simple sentences from. Then, we create three types of sentences, validated to be correct and convey the same meaning in both orders by an in-house annotator who is a native German speaker. Ones with both marked such as: {\it den Ball bringt der Hund} (lit. the dog brings the ball), ones with only the subject marked: 
{\it das Pfred drängt der Hund} (the dog urges the horse), and ones with only the object.

The three lists of sentences are:
\begin{itemize}
\item
Den \{Ball, Stein, Tisch, Hamster\} \{bringt, wirft, drückt\} \{das Kind, die Mutter, das Mädchen\}
\item
Das \{Pferd, Kind, Mädchen\} \{drängt, drückt, zieht\} der \{Vater, Hund, Student\}
\item
Den \{Ball, Stein, Tisch, Hamster\} \{bringt, wirft, drückt\} der \{Vater, Hund, Student\}
\end{itemize}

Then, we switch the object and subject and calculate how often is the translation correct in terms of places. We disregard other errors such as choice of verb in English.

Interestingly, as seen in the results section~\S\ref{sec:res}, both networks are quite bad at it (although the syntactic variant is better).

\begin{table}[tbp]
\begin{tabular}{rrr}
\multicolumn{1}{l}{} & \multicolumn{1}{l}{Vanilla} & \multicolumn{1}{l}{{\sc Parent}} \\
Object & 6 & 6 \\
Subject & 5 & 8 \\
Both & 10 & 13
\end{tabular}
\caption{Amount of sentences where the rare order (OVS) in German was still well corrected. In rows, what had unambiguous casing.}
\end{table}

\section{Results with the Large}\label{sec:large}
We include the full results over the two larger models {\sc Parent} and the Vanilla. While overall results are comparable, {\sc Parent}  consistently performs better on the challenge sets, often with large margins.
\begin{table*}[tbp]
\small
\begin{tabular}{lllllllllllll}
 & \multicolumn{4}{c}{Preposition Stranding} & \multicolumn{4}{c}{Particle} & \multicolumn{4}{c}{Reflexive} \\
 & Books &  & News &  & Books &  & News &  & Books &  & News &  \\
 & BLEU & chrF+ & BLEU & chrF+ & BLEU & chrF+ & BLEU & chrF+ & BLEU & chrF+ & BLEU & chrF+ \\
Vanilla & \multicolumn{1}{r}{8.70} & \multicolumn{1}{r}{33.58} & \multicolumn{1}{r}{\textbf{13.82}} & \multicolumn{1}{r}{43.41} & \multicolumn{1}{r}{8.59} & \multicolumn{1}{r}{32.66} & \multicolumn{1}{r}{\textbf{15.28}} & \multicolumn{1}{r}{44.28} & \multicolumn{1}{r}{8.54} & \multicolumn{1}{r}{32.85} & \multicolumn{1}{r}{18.90} & \multicolumn{1}{r}{45.82} \\
{\sc Parent} & \multicolumn{1}{r}{\textbf{9.03}} & \multicolumn{1}{r}{\textbf{34.83}} & \multicolumn{1}{r}{11.53} & \multicolumn{1}{r}{\textbf{45.12}} & \multicolumn{1}{r}{\textbf{8.59}} & \multicolumn{1}{r}{\textbf{33.71}} & \multicolumn{1}{r}{14.99} & \multicolumn{1}{r}{\textbf{45.90}} & \multicolumn{1}{r}{\textbf{9.05}} & \multicolumn{1}{r}{\textbf{34.11}} & \multicolumn{1}{r}{\textbf{20.79}} & \multicolumn{1}{r}{\textbf{46.73}}
\end{tabular}
\caption{Source challenge sets for En-De translation of large models. {\sc Parent} outperforms the Vanilla. \label{ap:tab:src_large_syn}}
\end{table*}

\begin{table*}[tbp]
\small
\centering
\begin{tabular}{lrlrlrlll}
 & 2013 &  & 2014 &  & 2015 &  & Average &  \\
 & \multicolumn{1}{l}{BLEU} & chrF+ & \multicolumn{1}{l}{BLEU} & chrF+ & \multicolumn{1}{l}{BLEU} & chrF+ & BLEU & chrF+ \\
Vanilla & \textbf{23.64} & \multicolumn{1}{r}{53.44} & 21.94 & \multicolumn{1}{r}{53.13} & \textbf{21.60} & \multicolumn{1}{r}{\textbf{50.84}} & \multicolumn{1}{r}{22.39} & \multicolumn{1}{r}{52.47} \\
{\sc Parent} & 23.56 & \multicolumn{1}{r}{\textbf{54.08}} & \textbf{22.11} & \multicolumn{1}{r}{\textbf{53.77}} & 20.69 & \multicolumn{1}{r}{49.16} & \multicolumn{1}{r}{22.12} & \multicolumn{1}{r}{52.34}
\end{tabular}
\caption{Test sets for En-De translation of large models. \label{ap:tab:large}}
\end{table*}

\begin{table*}[tbp]
\small
\centering
\begin{tabular}{lllllllll}
 & \multicolumn{4}{c}{Particle} & \multicolumn{4}{c}{Reflexive} \\
 & Books &  & News &  & Books &  & News &  \\
 & BLEU & chrF+ & BLEU & chrF+ & BLEU & chrF+ & BLEU & chrF+ \\
Vanilla & \multicolumn{1}{r}{4.14} & \multicolumn{1}{r}{20.72} & \multicolumn{1}{r}{20.31} & \multicolumn{1}{r}{49.04} & \multicolumn{1}{r}{8.08} & \multicolumn{1}{r}{32.38} & \multicolumn{1}{r}{20.65} & \multicolumn{1}{r}{49.09} \\
{\sc Parent} & \multicolumn{1}{r}{\textbf{8.37}} & \multicolumn{1}{r}{\textbf{33.78}} & \multicolumn{1}{r}{\textbf{20.54}} & \multicolumn{1}{r}{\textbf{49.99}} & \multicolumn{1}{r}{\textbf{8.60}} & \multicolumn{1}{r}{\textbf{33.49}} & \multicolumn{1}{r}{\textbf{21.39}} & \multicolumn{1}{r}{\textbf{50.01}}
\end{tabular}
\caption{Target challenge sets for En-De translation of large models. {\sc Parent} outperforms the Vanilla. \label{ap:trg_large_syn}}
\end{table*}
\FloatBarrier
\section{Additional Results}\label{sec:res}
We include here the full results including ablations that were omitted in the paper due to space considerations. For ease of comparison we also split them by challenge direction (source Table \ref{tab:src_challenge} and target Table \ref{tab:trg_all_challenge}).
Note that improvements in the syntactic aspect could also be seen in the ablations (not reported in the main paper). Moreover, BiTran improves over the Vanilla even as a standalone architecture.

\begin{table*}[tbh!]
\centering
\begin{subtable}{\textwidth}
\centering\begin{tabular}{lrrrrrrrr}
 & \multicolumn{4}{c}{Particle} & \multicolumn{4}{c}{Reflexive} \\
 & \multicolumn{2}{c}{Books} & \multicolumn{2}{c}{News}  & \multicolumn{2}{c}{Books}  & \multicolumn{2}{c}{News}  \\
 & \multicolumn{1}{l}{BLEU} & \multicolumn{1}{l}{chrF+} & \multicolumn{1}{l}{BLEU} & \multicolumn{1}{l}{chrF+} & \multicolumn{1}{l}{BLEU} & \multicolumn{1}{l}{chrF+} & \multicolumn{1}{l}{BLEU} & \multicolumn{1}{l}{chrF+} \\
Vanilla & 7.15 & 27.66 & 17.79 & 44.91 & 6.83 & 26.84 & 19.68 & 45.06 \\
{\sc Parent} & \textbf{7.82} & 28.43 & 19.66 & 46.32 & \textbf{7.49} & 27.70 & \textbf{20.97} & 47.07 \\
GCN & 7.32 & 27.67 & \textbf{20.13} & \textbf{46.77} & 7.11 & 27.16 & 20.68 & 47.15 \\
\addlinespace[0.4cm]
\midrule
BiTrans & 7.02 & 27.60 & 18.58 & 45.09 & 6.8 & 26.90 & 19.87 & 45.89 \\
Linearized & 7.44 & 28.05 & 19.2 & 46.21 & 7.27 & 27.43 & 20.25 & 46.92 \\
- Gates & 7.62 & 28.23 & 19.71 & 46.36 & 7.38 & 27.65 & 20.74 & 47.19 \\
- Labels & 7.75 & \textbf{28.60} & 19.01 & 46.51 & 7.44 & \textbf{27.90} & 20.81 & \textbf{47.32}  
\end{tabular}
\caption{Syntactic source challenge sets for De-En}
\label{tab:en_de_challenge}
\vspace{.3cm}
\end{subtable}
\begin{subtable}[h]{\textwidth}
\resizebox{\textwidth}{!}{%
\begin{tabular}{lrrrrrrrrrrrr}
 & \multicolumn{4}{c}{Preposition Stranding} & \multicolumn{4}{c}{Particle} & \multicolumn{4}{c}{Reflexive} \\
 & \multicolumn{2}{c}{Books}  & \multicolumn{2}{c}{News}  & \multicolumn{2}{c}{Books}  & \multicolumn{2}{c}{News}  & \multicolumn{2}{c}{Books}  & \multicolumn{2}{c}{News}  \\
 & \multicolumn{1}{l}{BLEU} & \multicolumn{1}{l}{chrF+} & \multicolumn{1}{l}{BLEU} & \multicolumn{1}{l}{chrF+} & \multicolumn{1}{l}{BLEU} & \multicolumn{1}{l}{chrF+} & \multicolumn{1}{l}{BLEU} & \multicolumn{1}{l}{chrF+} & \multicolumn{1}{l}{BLEU} & \multicolumn{1}{l}{chrF+} & \multicolumn{1}{l}{BLEU} & \multicolumn{1}{l}{chrF+} \\
Vanilla & 5.95 & 25.88 & 9.96 & 36.96 & 5.37 & 24.69 & 9.39 & 39.19 & 5.32 & 24.71 & 16.48 & 42.04 \\
{\sc Parent} & \textbf{6.21} & \textbf{28.12} & 11.17 & \textbf{41.13} & 5.47 & 25.74 & 11.93 & 41.24 & 5.71 & \textbf{26.22} & 15.56 & 42.76 \\
GCN & \textbf{6.21} & 27.27 & 11.31 & 40.48 & 5.51 & 25.53 & 10.35 & 39.83 & 5.46 & 25.70 & 16.45 & \textbf{43.03} \\
\addlinespace[0.4cm] \midrule
BiTrans & 5.3 & 26.38 & 10.56 & 38.05 & \textbf{6.07} & \textbf{26.08} & 10.21 & 39.48 & \textbf{5.77} & 26.01 & 13.91 & 37.74 \\ 
Linearized & 5.99 & 26.90 & 8.86 & 39.12 & 5.24 & 25.42 & 10.45 & 39.56 & 5.48 & 25.47 & 14.94 & 42.15 \\
- Gates & 5.29 & 25.86 & \textbf{11.64} & 40.51 & 5.3 & 25.03 & 10.01 & 38.64 & 5.31 & 25.41 & 12.08 & 37.00 \\
- Labels & 5.83 & 27.05 & 8.62 & 38.33 & 5.41 & 25.62 & \textbf{11.98} & \textbf{41.79} & 5.42 & 25.67 & \textbf{16.55} & 41.65 \\
\end{tabular}%
}
\label{ap:tab:de_en_challenge}
\caption{Syntactic source challenge sets for En-De}
\end{subtable}
\caption{Results on the syntactic challenge sets, both on the large challenges from book domain and the smaller ones from news. Models include Vanilla and Bidirectional Transformer baselines (top) and the GCN and {\sc Parent} syntactic variants (middle). Ablated models (bottom) include Vanilla with linearized syntax (Linearized), GCN without labels or gating (-Gates) and GCN without labels (-Labels).
Among the baselines, BiTrans is better. It is inconclusive which syntactic method is best, but they are significantly superior to both baselines.\label{tab:src_challenge}}
\end{table*}

\begin{table*}[tbh!]
\centering
\begin{subtable}{\textwidth}
\centering
\resizebox{\textwidth}{!}{%
\begin{tabular}{lrrrrrrrrrrrr}
 & \multicolumn{4}{c}{Preposition Stranding} & \multicolumn{4}{c}{Particle} & \multicolumn{4}{c}{Reflexive} \\
 & \multicolumn{1}{l}{Books} & \multicolumn{1}{l}{} & \multicolumn{1}{l}{News} & \multicolumn{1}{l}{} & \multicolumn{1}{l}{Books} & \multicolumn{1}{l}{} & \multicolumn{1}{l}{News} & \multicolumn{1}{l}{} & \multicolumn{1}{l}{Books} & \multicolumn{1}{l}{} & \multicolumn{1}{l}{News} & \multicolumn{1}{l}{} \\
 & \multicolumn{1}{l}{BLEU} & \multicolumn{1}{l}{chrF+} & \multicolumn{1}{l}{BLEU} & \multicolumn{1}{l}{chrF+} & \multicolumn{1}{l}{BLEU} & \multicolumn{1}{l}{chrF+} & \multicolumn{1}{l}{BLEU} & \multicolumn{1}{l}{chrF+} & \multicolumn{1}{l}{BLEU} & \multicolumn{1}{l}{chrF+} & \multicolumn{1}{l}{BLEU} & \multicolumn{1}{l}{chrF+} \\
Vanilla & 6.38 & 27.30 & 9.18 & 38.22 & 6.53 & 25.70 & 10.54 & 38.28 & 6.15 & 25.94 & 17.2 & 43.12 \\
{\sc Parent} & \textbf{7.59} & \textbf{27.87} & \textbf{10.81} & 39.22 & \textbf{7.07} & \textbf{26.50} & 9.72 & 39.57 & \textbf{6.82} & 26.58 & 17.56 & 44.00 \\
GCN & 6.33 & 26.60 & 10.14 & \textbf{41.00} & 6.69 & 26.16 & 10.6 & 39.81 & 6.33 & 25.83 & \textbf{20.16} & 44.19 \\
\addlinespace[0.4cm] \midrule
BiTrans & 6.75 & 27.44 & 8.92 & 37.76 & 6.29 & 25.69 & 10.77 & 39.15 & 6.24 & 25.93 & 17.22 & 43.96 \\
Linearized & 6.79 & 27.46 & 7.79 & 39.62 & 6.55 & 25.96 & \textbf{12.95} & \textbf{40.78} & 6.56 & 26.28 & 16.38 & 43.76 \\
 - Gates & 6.89 & 27.31 & 10.46 & 40.80 & 6.53 & 26.26 & 12.45 & 40.70 & 6.62 & 26.50 & 15.97 & 43.10\\
- Labels & 7.05 & 27.51 & 9.89 & 38.24 & 6.98 & 26.42 & 12.83 & 40.18 & 6.62 & \textbf{26.65} & 18.9 & \textbf{46.59} 
\end{tabular}%
}
\caption{Syntactic target challenge sets for De-En}
\label{ap:tab:trg_en_de_challenge}
\vspace{.3cm}
\end{subtable}
\begin{subtable}[h]{\textwidth}
\centering
\begin{tabular}{lrrrrrrrr}
 & \multicolumn{4}{c}{Particle} & \multicolumn{4}{c}{Reflexive} \\
 & \multicolumn{1}{l}{Books} & \multicolumn{1}{l}{} & \multicolumn{1}{l}{News} & \multicolumn{1}{l}{} & \multicolumn{1}{l}{Books} & \multicolumn{1}{l}{} & \multicolumn{1}{l}{News} & \multicolumn{1}{l}{} \\
 & \multicolumn{1}{l}{BLEU} & \multicolumn{1}{l}{chrF+} & \multicolumn{1}{l}{BLEU} & \multicolumn{1}{l}{chrF+} & \multicolumn{1}{l}{BLEU} & \multicolumn{1}{l}{chrF+} & \multicolumn{1}{l}{BLEU} & \multicolumn{1}{l}{chrF+} \\
Vanilla & 5.4 & 25.84 & \textbf{16.24} & 43.22 & 5.12 & 24.94 & 16.47 & 42.71 \\
{\sc Parent} & 5.52 & \textbf{26.96} & 16.19 & \textbf{44.83} & 5.37 & \textbf{26.31} & \textbf{16.86} & \textbf{44.30} \\
GCN & 5.6 & 26.74 & 15.57 & 43.23 & 5.34 & 25.91 & 16.44 & 43.52 \\
\addlinespace[0.4cm] \midrule
BiTrans & \textbf{5.81} & 26.79 & 15.84 & 43.25 & \textbf{5.43} & 25.88 & 16.33 & 42.44 \\
+ Linearized & 5.32 & 26.30 & 15.69 & 43.77 & 5.07 & 25.57 & 16.19 & 43.07 \\
- Gates & 5.31 & 26.21 & 15.49 & 43.45 & 5.01 & 25.30 & 15.67 & 43.13 \\
- Labels & 5.56 & 26.55 & 15.78 & 43.96 & 5.24 & 25.67 & 16.8 & 43.65
\end{tabular}%

\label{tab:trg_de_en_challenge}
\caption{Syntactic target challenge sets for En-De}
\end{subtable}
\caption{Results on the syntactic challenge sets, both on the large challenges from book domain and the smaller ones from news. Models include Vanilla and Bidirectional Transformer baselines (top) and the GCN and {\sc Parent} syntactic variants (middle). Ablated models (bottom) include the Vanilla with linearized syntax (Linearized), GCN without labels or gating (-Gates) and GCN without labels (-Labels).
Among the baselines, BiTrans is better. It is inconclusive which syntactic method is best, but they are significantly superior to both baselines.\label{tab:trg_all_challenge}}
\end{table*}

\begin{table*}[tbh!]
\centering
\begin{subtable}{\textwidth}
\centering
\begin{tabular}{lrr|rr|rr|rr}
 & \multicolumn{2}{c}{2013} & \multicolumn{2}{c}{2014} & \multicolumn{2}{c}{2015} & \multicolumn{2}{c}{Average} \\
 & \multicolumn{1}{c}{BLEU} & \multicolumn{1}{c}{chrF+} & \multicolumn{1}{c}{BLEU} & \multicolumn{1}{c}{chrF+} & \multicolumn{1}{c}{BLEU} & \multicolumn{1}{c}{chrF+} & \multicolumn{1}{c}{BLEU} & \multicolumn{1}{c}{chrF+} \\
Vanilla & 17.61 & 45.54 & 18.23 & 47.29 & 19.57 & 47.50 & 18.47 & 46.78 \\
BiTrans & 17.64 & 45.66 & 18.34 & 47.53 & 19.33 & 47.61 & 18.44 & 46.93 \\ \midrule
{\sc Parent} & \textbf{18.11} & \textbf{46.75} & 18.6 & \textbf{48.46} & \textbf{20.55} & \textbf{49.20} & \textbf{19.09} & \textbf{48.14} \\
GCN & 18.03 & 46.43 & \textbf{18.86} & \textbf{48.46} & 20.32 & 48.90 & 19.07 & 47.93 \\
\addlinespace[0.4cm] \midrule
Linearized & 17.71 & 46.07 & 18.39 & 47.69 & 19.81 & 48.36 & 18.64 & 47.37 \\
- Gates & 17.81 & 46.12 & 18.43 & 48.08 & 20.06 & 48.62 & 18.77 & 47.61 \\
- Labels & 17.98 & 46.40 & 18.77 & 48.29 & 19.96 & 48.73 & 18.90 & 47.80 \\
\end{tabular}%
\caption{Test sets for En-De translation}
\end{subtable}
\begin{subtable}{\textwidth}
\centering
\begin{tabular}{lrr|rr|rr|rr}
 & \multicolumn{2}{c}{2013} & \multicolumn{2}{c}{2014} & \multicolumn{2}{c}{2015} & \multicolumn{2}{c}{Average} \\
 & \multicolumn{1}{c}{BLEU} & \multicolumn{1}{c}{chrF+} & \multicolumn{1}{c}{BLEU} & \multicolumn{1}{c}{chrF+} & \multicolumn{1}{c}{BLEU} & \multicolumn{1}{c}{chrF+} & \multicolumn{1}{c}{BLEU} & \multicolumn{1}{c}{chrF+} \\
Vanilla & 21.51 & 48.20 & 21.40 & 48.46 & 21.44 & 48.13 & 21.45 & 48.26 \\
BiTrans & 21.63 & 48.48 & 21.42 & 48.86 & 21.38 & 48.54 & 21.48 & 48.63 \\ \midrule
{\sc Parent} & \textbf{22.46} & 49.24 & 21.75 & 49.41 & 22.14 & 49.31 & 22.12 & 49.32 \\
GCN & 22.33 & 49.27 & 21.76 & 49.71 & \textbf{22.43} & \textbf{49.73} & \textbf{22.17} & 49.57 \\ 
\addlinespace[0.4cm] \midrule
Linearized & 21.95 & 49.27 & 21.83 & \textbf{49.79} & 22.20 & 49.70 & 21.99 & \textbf{49.59} \\
- Gates & 22.28 & 49.33 & \textbf{21.89} & 49.68 & 22.04 & 49.39 & 22.07 & 49.46 \\
- Labels & 22.21 & \textbf{49.46} & 21.75 & 49.73 & 22.26 & 49.57 & 22.07 & \textbf{49.59} \\
\end{tabular}%
\caption{Test sets for De-En translation}
\end{subtable}
\caption{ En-De and De-En results on newstest 2013-15. Ablated models include the Transformer decoder with linearized syntax (Linearized), GCN without labels or gating (-Gates) and GCN without labels (-Labels). The syntactic variants consistently outperfom the vanilla and ablated variants, and the Bidirectional Transformer (BiTrans) slightly outperforms Vanilla Transformer.\label{ap:tab:english-german}}
\end{table*}
\FloatBarrier
\subsection{Noisy data}\label{sec:full_noisy}
Table \ref{ap:tab:rus} presents the two tables side by side for ease of comparison. The one on larger noisy Russian train set and the cleaner one.

\begin{table*}[tbh!]
\small
\centering
\begin{subtable}{\textwidth}
\centering
\begin{tabular}{lrrrrrrrr}
 & \multicolumn{2}{c}{2013} & \multicolumn{2}{c}{2014} & \multicolumn{2}{c}{2015} & \multicolumn{2}{c}{Average} \\
 & \multicolumn{1}{c}{BLEU} & \multicolumn{1}{c}{chrF+} & \multicolumn{1}{c}{BLEU} & \multicolumn{1}{c}{chrF+} & \multicolumn{1}{c}{BLEU} & \multicolumn{1}{c}{chrF+} & \multicolumn{1}{c}{BLEU} & \multicolumn{1}{c}{chrF+} \\
Vanilla & 13.20 & 38.72 & 17.17 & 43.69 & 14.19 & 40.87 & 14.85 & 41.09 \\
BiTran & 13.13 & 39.10 & 17.63 & 44.63 & 14.59 & 41.52 & 15.12 & 41.75 \\
GCN & 13.25 & 40.31 & 17.86 & 46.09 & 15.38 & 43.09 & 15.50 & 43.16 \\
{\sc Parent} & \textbf{13.61} & \textbf{40.67} & \textbf{18.53} & \textbf{46.44} & \textbf{15.75} & \textbf{43.57} & \textbf{15.96} & \textbf{43.56}
\end{tabular}%
\caption{Test sets for En-Ru translation trained on news data\label{ap:tab:rus}}
\end{subtable}
\begin{subtable}{\textwidth}
\centering
\begin{tabular}{@{}lrrrrrrrr@{}}
 & \multicolumn{2}{c}{2013} & \multicolumn{2}{c}{2014} & \multicolumn{2}{c}{2015} & \multicolumn{2}{c}{Average} \\
 & \multicolumn{1}{c}{BLEU} & \multicolumn{1}{c}{chrF+} & \multicolumn{1}{c}{BLEU} & \multicolumn{1}{c}{chrF+} & \multicolumn{1}{c}{BLEU} & \multicolumn{1}{c}{chrF+} & \multicolumn{1}{c}{BLEU} & \multicolumn{1}{c}{chrF+} \\
Vanilla & 16.84 & 44.28 & 20.12 & 47.7 & 14.74 & 40.92 & 17.23 & 44.30 \\
BiTran & 16.84 & 44.46 & \textbf{20.61} & 48.17 & \textbf{14.79} & 41.05 & \textbf{17.41} & 44.56 \\
GCN & \textbf{17.11} & \textbf{45.55} & 20.29 & 48.67 & 14.6 & 41.63 & 17.33 & 45.28 \\
{\sc Parent} & 16.8 & 45.42 & 20.2 & \textbf{48.95} & 14.59 & \textbf{41.73} & 17.20 & \textbf{45.37}
\end{tabular}%
\caption{Test sets for En-Ru translation trained on noisy data}
\end{subtable}
\caption{En-Ru results on newstest 2013-15 trained on clean (top) or noisy (bottom) data. Models include Vanilla, Bidirectional Transformer and syntactic variants. The syntactic ones improve over all datasets and on average.\label{ap:tab:rus_wmt}}
\end{table*}

\end{document}